\PassOptionsToPackage{dvipsnames}{xcolor}

\documentclass[twoside]{article}

\usepackage[accepted]{aistats2024}
\usepackage{apacite}

\usepackage{graphicx}
\usepackage[hyphens]{url}

\usepackage[colorinlistoftodos,prependcaption,textsize=tiny]{todonotes}

\usepackage[utf8]{inputenc}
\usepackage[T1]{fontenc}    
\usepackage{hyperref}       
\usepackage{url}            
\usepackage{booktabs}       
\usepackage{amsfonts}       
\usepackage{nicefrac}       
\usepackage{microtype}      
\usepackage{xcolor}         
\usepackage{multicol}
\usepackage{array}
\usepackage{amsmath, amsthm, amssymb, amsfonts}
\usepackage{mathtools}
\usepackage{bm}
\usepackage{enumitem}
\usepackage{float}
\usepackage{tcolorbox}
\usepackage{tabularx} 
\usepackage{soul}
\usepackage{capt-of}
\usepackage{doi}
\usepackage{algorithm}
\usepackage{algpseudocode}
\usepackage{xspace}
\usepackage{wrapfig,graphbox}
\usepackage[capitalise,nameinlink]{cleveref}
\usepackage{relsize}
\usepackage{caption}
\usepackage{subcaption}

\newcommand{\IC}{\textsc{IC}\xspace}
\newcommand{\ICs}{\textsc{ICs}\xspace}
\newcommand{\PIC}{\textsc{PIC}\xspace}
\newcommand{\PICs}{\textsc{PICs}\xspace}
\newcommand{\QPC}{\textsc{QPC}\xspace}
\newcommand{\QPCs}{\textsc{QPCs}\xspace}

\DeclareFontFamily{U}{mathx}{}
\DeclareFontShape{U}{mathx}{m}{n}{<-> mathx10}{}
\DeclareSymbolFont{mathx}{U}{mathx}{m}{n}
\DeclareMathAccent{\widecheck}{0}{mathx}{"71}

\newcommand*\circled[1]{\tikz[baseline=(char.base)]{
  \node[shape=circle,draw,inner sep=0.10pt] (char) {#1};}}

\newcommand{\pa}{\mathsf{pa}}

\NewDocumentCommand{\intz}{o}{
  \IfValueTF{#1}{%
    \tilde{z}_{#1}%
  }{%
    \tilde{\rvz}%
  }%
}
\NewDocumentCommand{\intw}{o}{
  \IfValueTF{#1}{%
    \tilde{w}_{#1}%
  }{%
    \tilde{\rvw}%
  }%
}

\newcommand{\intu}[1][i]{\circled{$\smallint\!\!z_{#1}$}}

\newcommand{\sumnet}{\gS_{\phi(i)}}         
\newcommand{\innet}{\gI_{\phi(i)}}           
\newcommand{\sumnets}{\gS_{\phi}}           
\newcommand{\innets}{\gI_{\phi}}             

\newcommand{\inscope}{\ensuremath{\textsc{in}}}

\newsavebox{\inputunitbox}
\savebox{\inputunitbox}{%
	\tikz[x=.5em,y=1ex,baseline=0]{%
		\draw[line width=.5pt] plot[domain=-.66:.66] (\x,{1.2*exp(-14*\x*\x)});
		\draw[line width=.5pt] (0,.625ex) circle[radius=1.15ex];
	}%
}
\newcommand{\inputunit}{\usebox\inputunitbox\xspace}

\newcommand{\X}{\mathbf{X}}
\newcommand{\x}{\mathbf{x}}

\newcommand{\Z}{\mathbf{Z}}

\newcommand{\cbar}{\,|\,}

\newcommand{\node}{\mathsf{N}}

\def\eqref#1{equation~\ref{#1}}

\def\1{\bm{1}}

\def\rvb{{\mathbf{b}}}

\def\rvf{{\mathbf{f}}}

\def\rvw{{\mathbf{w}}}
\def\rvx{{\mathbf{x}}}

\def\rvz{{\mathbf{z}}}

\def\rmA{{\mathbf{A}}}

\def\rmX{{\mathbf{X}}}

\def\rmZ{{\mathbf{Z}}}

\DeclareMathAlphabet{\mathsfit}{\encodingdefault}{\sfdefault}{m}{sl}
\SetMathAlphabet{\mathsfit}{bold}{\encodingdefault}{\sfdefault}{bx}{n}

\def\gC{{\mathcal{C}}}

\def\gE{{\mathcal{E}}}

\def\gG{{\mathcal{G}}}

\def\gI{{\mathcal{I}}}

\def\gN{{\mathcal{N}}}
\def\gO{{\mathcal{O}}}
\def\gP{{\mathcal{P}}}

\def\gS{{\mathcal{S}}}

\def\sR{{\mathbb{R}}}

\newcommand{\E}{\mathbb{E}}

\usepackage[round]{natbib}

\algdef{SE}[FUNCTION]{Function}{EndFunction}%
   [2]{\algorithmicfunction\ \textproc{#1}\ifthenelse{\equal{#2}{}}{}{(#2)}}%
   {\algorithmicend\ \algorithmicfunction}%

\hypersetup{
colorlinks=true,linkcolor=RedViolet,citecolor=teal
}

\definecolor{tomato}{HTML}{E53935}
\begin{document}

%

%
\runningauthor{Gala, de Campos, Peharz, Vergari, Quaeghebeur}

\twocolumn[

\aistatstitle{Probabilistic Integral Circuits}
\aistatsauthor{
\href{mailto: Gennaro Gala <g.gala@tue.nl>}{Gennaro Gala}\textsuperscript{\rm 1}\space\space
Cassio de Campos\textsuperscript{\rm 1}\space\space
Robert Peharz\textsuperscript{\rm 1,2}\space\space
Antonio Vergari\textsuperscript{\rm 3}\space\space
Erik Quaeghebeur\textsuperscript{\rm 1}
}
\vspace{2mm}
\aistatsaddress{
\textsuperscript{\rm 1}Eindhoven University of Technology, NL\\
\textsuperscript{\rm 2}Graz University of Technology, AT\\
\textsuperscript{\rm 3}School of Informatics, University of Edinburgh, UK}
]

\begin{abstract}
Continuous latent variables (LVs) are a key ingredient of many generative models, as they allow modelling expressive mixtures with an uncountable number of components.
In contrast, probabilistic circuits (PCs) are hierarchical discrete mixtures represented as computational graphs composed of input, sum and product units.
Unlike continuous LV models, PCs provide tractable inference but are limited to discrete LVs with categorical (i.e.~unordered) states.
We bridge these model classes by introducing \emph{probabilistic integral circuits} (\PICs), a new language of computational graphs that extends PCs with integral units representing continuous LVs.
In the first place, 
\PICs are \emph{symbolic} computational graphs and are fully tractable in simple cases where analytical integration is possible.
In practice, we parameterise \PICs with light-weight neural nets delivering an intractable hierarchical continuous mixture that can be approximated arbitrarily well with large PCs using numerical quadrature.
On several distribution estimation benchmarks, we show that such \PIC-approximating PCs systematically outperform PCs commonly learned via expectation-maximization or SGD.
\end{abstract}

\vspace{-4.145mm}
\section{Introduction}

\begin{figure*}[!ht]
\centering
\includegraphics[scale=0.54]{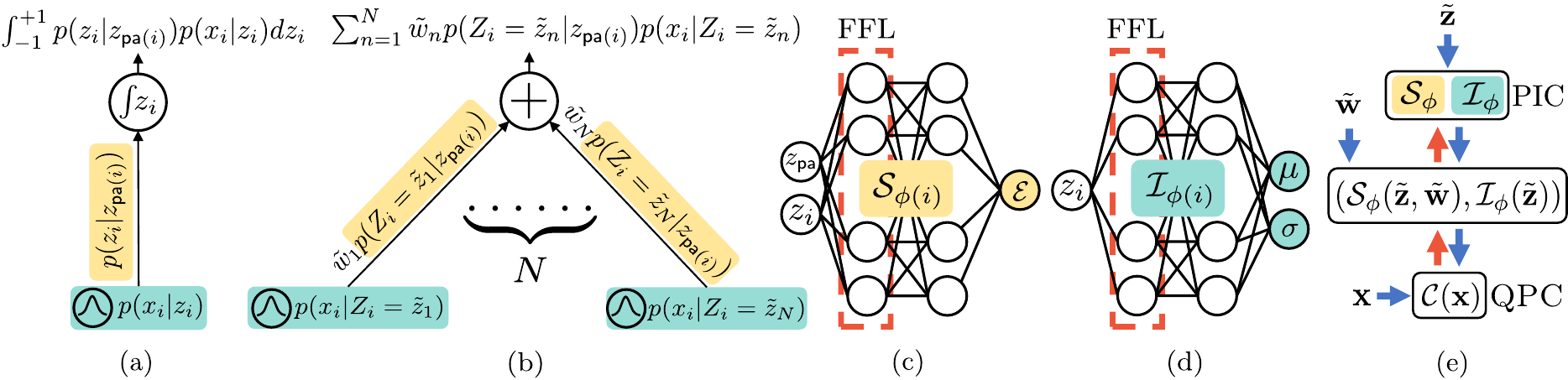}
\caption{
\textbf{Our contributions.}
We extend PCs with symbolic integral units, leading to so-called PICs (\cref{sec:pic}) (a) which we approximate via numerical quadrature as discrete sum units (b).
We use neural nets (c) and (d) to parameterise $p(Z_i \cbar Z_{\pa(i)})$ and $p(X_i \cbar Z_i)$ respectively.
In (e), we sketch the training computation graph: We materialise \QPC parameters $(\sumnets(\intz, \intw), \innets(\intz))$ in the forward pass (blue arrows) w.r.t.\ quadrature points $\intz$ and weights $\intw$, and propagate gradients in the backward 
pass (orange arrows) to maximise \QPC log-likelihood $\gC(\rvx)$.
}
\label{fig:quad}
\end{figure*}

Many successful probabilistic models build upon \emph{continuous mixtures} (CMs) \citep{bond2021deep}, which model the data-generating distribution as:
\begin{gather}
\textstyle
p(\rmX) = \E_{p(\rmZ)} \left [ p(\rmX \cbar \rmZ ) \right ] = \int p(\rvz) \, p(\rmX \cbar \rvz ) \, \mathrm{d}\rvz,
\label{eq:continuous_mixture}
\end{gather}
where $p(\rmZ)$ is a mixing distribution over \emph{continuous} latent variables (LVs) $\rmZ$, $p(\rmX \cbar \rmZ)$ is a conditional distribution defining mixture components, and $p(\rmX)$ is the distribution over observables $\rmX$ obtained by marginalising out $\rmZ$ from the joint  $p(\rmX, \rmZ) = p(\rmX \cbar \rmZ)\,p(\rmZ)$.

CMs underpin the success of deep generative models like variational autoencoders (VAEs) \citep{kingma2013auto} and generative adversarial networks \citep{goodfellow2014generative}.
%
%
These models are designed such that (i)~$p(\rmZ)$ has a simple parametrisation, e.g., an isotropic Gaussian, and (ii)~the mixture components $p(\rmX \cbar \rmZ)$ are parameterised by neural networks taking $\rmZ$ as input.
%
%
However, the ability of CMs to support tractable probabilistic inference is generally limited.
Specifically, \emph{marginalisation} and \emph{conditioning}---which together form a consistent reasoning process \citep{Ghahramani2015, Jaynes2003}---are largely intractable in these models, mainly due to the generally high-dimensional integral in \autoref{eq:continuous_mixture}.

In contrast, \emph{Probabilistic Circuits} (PCs) \citep{vergari2019tractable, choiprobabilistic} represent a unified framework for tractable probabilistic models, providing a wide range of exact and efficient inference routines.
PCs are computational graphs composed of input, sum and product units, and they represent hierarchical \emph{discrete} mixtures (DMs).
This discrete nature is crucial for their tractability, as it allows tractable marginalization of LVs in a single feedforward pass.
However, although PCs are deep and structured models encoding an exponential number of mixture components, even a shallow CM has been recently shown to often outperform them \citep{correia2023continuous}.
It is indeed speculated that DMs---like PCs---are hard to learn, while shallow CMs---like VAEs---are easier to learn or generalise better \citep{liu2023scaling}.
For instance, applying Gaussian Mixture Models, a special case of PCs, to complex, high-dimensional data presents many challenges, such as dealing with sensitivity to initialization \citep{bender23continuously}.

In this paper, we show how to incorporate the benefits of CMs into tractable probabilistic models by extending the language of PCs with continuous LVs, which we address with numerical integration.
Specifically, we make the following contributions (which are visually summarized in \autoref{fig:quad}):
\begin{itemize}[leftmargin=*, wide=0pt, topsep=0pt, itemsep=1pt]

    \item We introduce \emph{probabilistic integral circuits} (\PICs), a new language of computational graphs that extends PCs with integral units, allowing them to model continuous LVs.
    Different from PCs, they represent \emph{symbolic} computational graphs and allow tractable inference when analytical integration is possible (\cref{sec:pic}).

    \item If analytic solutions are not available, e.g., for PICs encoding neural CMs, we design and apply inference routines based on numerical quadrature, allowing to approximate PICs arbitrary well.
    We call the output of this process \QPC (\cref{sec:qpc}), a PC encoding the \textbf{q}uadrature of a deep \PIC, thus generalising ideas of \citet{correia2023continuous}.
    
    \item We show how to parameterise \PICs with light-weight energy-based models, delivering a flexible generative model that we train via maximum-likelihood using quadrature rules (\cref{sec:learnpic}). 
    
    \item On several distribution estimation benchmarks (\cref{sec:exp}), \QPCs systematically outperform traditional PCs commonly learned via EM or SGD.
    
\end{itemize}

\section{Preliminaries}
\label{sec:background}


Before introducing \PICs, we recall how to represent hierarchies of LVs as graphical models and review the necessary background on circuits.
We conclude with a brief overview of numerical quadrature rules, as these will allow approximate inference in \PICs.

\subsection{Graphical models for CMs}
\label{sec:bn}

A hierarchy over random variables (RVs) $\mathbf{Y} = \{ Y_i \}_{i=1}^{D}$ can be represented as a Bayesian network (BN), encoded as a 
DAG with RVs as nodes and 
parameterized by a set of conditional distributions $p(Y_i \cbar \pa(Y_i))$,
where $\pa(Y_i)$ is the parent set of $Y_i$.
A BN encodes the joint $p(\mathbf{Y})=\prod_{i=1}^{D} p(Y_i \cbar \pa(Y_i))$.
We will consider some variables $\mathbf{Z}\subset\mathbf{Y}$ to be latent and the remaining variables $\mathbf{X}=\mathbf{Y}\setminus\mathbf{Z}$ observable.

\begin{wrapfigure}{r}{0.19\textwidth}
\vspace{-3.5mm}
\centering
\includegraphics[scale=0.45]{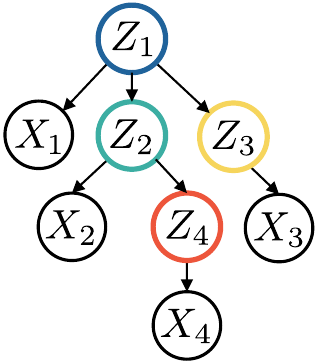}
\vspace{-2mm}
\caption{Latent Tree}
\label{fig:lt}
\vspace{-3.5mm}
\end{wrapfigure} 

\paragraph{LTMs}
Marginalizing LVs is a \#P-hard task \citep{DECAMPOS2020104627}, but becomes tractable for discrete BNs when the DAG has bounded treewidth, in particular when it is a tree, yielding the class of \textit{latent tree models} (LTMs, \autoref{fig:lt}) \citep{choi2011learning}.
However, continuous LTMs allow tractable marginalisation only if the parametric representations of the conditional probabilities allow for analytical integration, like in linear-Gaussian LTMs \citep{koller2009probabilistic}.

\subsection{Probabilistic Circuits}

\paragraph{Circuits}
A \emph{circuit} $\gC \equiv (\gG, \theta)$ is a computational graph $\gG$ parameterised by $\theta$ representing a function $\gC(\rmX)$ over its input variables $\rmX$.
$\gG$ has three computational unit types: \emph{input} \inputunit, \emph{sum} $\bigoplus$ and \emph{product} $\bigotimes$. 
An input unit $c$ represents a parameterisable function $f_c(\rmX_c)$, where $\rmX_c \subseteq \rmX$ is its \emph{scope}.
Any product or sum unit $c$ receives the outputs of its input units, denoted by the set $\inscope(c)$, and has as scope the union of the scopes of its inputs, $\X_c = \cup_{d \in \inscope(c)} \X_{d}$.
A product unit $c$ computes the product of its incoming inputs as $\prod_{d \in \inscope(c)} d(\X_d)$, whereas a sum unit $c$ computes the weighted sum $\sum_{d \in \inscope(c)} \theta_{d}^{c} \, d(\X_d)$.
Note that $\gC(\X)=c(\X)$ where $c$ is the circuit root unit.
We instantiate a sum or product unit $c$ as $\bigoplus([d]_{d \in \inscope(c)}, [\theta_{d}^{c}]_{d \in \inscope(c)})$ or $\bigotimes([d]_{d \in \inscope(c)})$ (cf.\ \cref{alg:pic2qpc}).

\paragraph{Probabilistic Circuits (PCs)}
A circuit $\gC$ is \emph{probabilistic} when $\gC(\rvx) \geq 0$ for any $\x$, i.e.\ $\gC(\X)$ is an unnormalized distribution.
For efficient renormalization it must satisfy two structural properties: \emph{smoothness} and \emph{decomposability}.
A PC is smooth when every sum unit has input units with the same scope, whereas is decomposable when every product unit has input units with disjoint scopes.
Furthermore, a PC is \emph{structured-decomposable}---a stricter form of decomposability---when any pair of product units sharing the same scope decomposes the same way, as in \autoref{fig:pic}(b) \citep{pipatsrisawat2008new, vergari2021compositional}.
PCs are hierarchical discrete mixtures \citep{peharz2016latent,zhao2016unified, trapp2019bayesian, yang2023bayesian}, i.e.~$\gC(\X)$ can be written as 
\begin{equation}
\textstyle
p(\X) = \sum_{\rvz} P(\rvz) \, p(\X \cbar \rvz),
\label{eq:discrete_mixture}
\end{equation}
where here $\rmZ$ is a \textbf{\emph{discrete}} latent vector, but otherwise the form is similar to the continuous mixture in \autoref{eq:continuous_mixture}.
The number of states of $\rmZ$, and thus the number of mixture components $p(\X \cbar \rvz)$, grows exponentially in the depth of the PC \citep{Peharz2015a, zhao2016unified}.

\paragraph{Discrete LV interpretation}
PCs can be compiled from discrete graphical models \citep{darwiche2009modeling, abramowitz1988handbook} or directly learned from data \citep{mari2023unifying, peharz2020random}.
In either case sum units can be interpreted as discrete (categorical) LVs \citep{peharz2016latent}.
As an example, consider the LTM in \autoref{fig:lt}.
If we assume its LVs to be categorical, it can be compiled into a smooth and structured-decomposable PC \citep{pmlr-v138-butz20a}.
Specifically, if we assume \emph{all} its LVs to have $N=3$ states, the compiled circuit would look like \autoref{fig:pic}(b), where \emph{regions} of $N=3$ units are arranged hierarchically and sum parameters encode conditional probability tables.

\subsection{Numerical Quadrature}
As shown by  \citet{correia2023continuous}, \emph{numerical quadrature rules} can be used to reduce inference in shallow CMs (\cref{eq:continuous_mixture}) to shallow DMs (\cref{eq:discrete_mixture}).
In this paper, we deal with \textit{deep} CMs via \textit{recursive} numerical quadrature. 

A {numerical quadrature rule} is an approximation of the definite integral of a function as a weighted sum of function evaluations at specified points within the domain of integration \citep{davis2007methods}.
Specifically, given some integrand $f:\sR \rightarrow \sR$ and interval $[a ,b]$, a quadrature rule consists of a set of $N$ integration points $\intz = (\intz[n])_{n=1}^N \in [a,b]^{N}$ and weights $\intw = (\intw[n])_{n=1}^{N} \in \sR^N$ which minimize the integration error $\varepsilon_N = \left|\smallint_{a}^{b} f(z) \, \mathrm{d}z - \sum_{n=1}^N \intw[n] f(\intz[n]) \right|$.
Such integration points define sub-intervals over which the integrand is approximated by polynomials. 
Note that the error $\varepsilon_N$ goes to zero as $N \rightarrow \infty$ and that $\sum_{n=1}^N\intw[n]=b-a$.

\paragraph{Types of quadratures}
\emph{Static} quadrature rules like midpoint, trapezoidal and Simpson's achieve error bounds of $\mathcal{O}(N^{-2})$, $\mathcal{O}(N^{-2})$ and $\mathcal{O}(N^{-4})$, respectively.
Their choice of $\intz$ \emph{exclusively} depends on the finite integration domain.
\emph{Gaussian} quadrature rules go a step further as they allow (i)~exact integration of any polynomial of degree up to $2N-1$ and (ii)~handling infinite integration domains. 
In particular, Gauss-Hermite quadrature allows to approximate expectations under a Gaussian-distributed variable, but the choice of $\intz$ depends on its $\mu$ and $\sigma$, i.e.\ it is integrand-dependent.
Finally, \emph{adaptive} quadratures approximate the integral of a function using static methods on adaptively refined sub-intervals, until an error tolerance is achieved.
The choice of $\intz$ is again integrand-dependent, as the refined sub-intervals are those where it is hard to approximate.

\section{Probabilistic Integral Circuits}
\label{sec:pic}

In this section, we introduce \emph{integral circuits} (\ICs), symbolic computational graphs leading to our main contribution, \emph{probabilistic integral circuits} (\PICs).
The inputs and outputs of every \IC unit are considered functions.
This is different from PCs, where marginalization of discrete LVs is explicitly
and exhaustively represented in the discrete circuit structure (cf. \cref{fig:quad}(a-b)).

\begin{figure*}[!ht]
\begin{minipage}[b]{.48\textwidth}
\begin{algorithm}[H]
\caption{\textbf{\PICs $\rightarrow$ \QPCs} via \emph{static} quadrature}
\label{alg:pic2qpc}
\begin{flushleft}
\textbf{Input} A tree-shaped \PIC $\gC \equiv (\gG, \Theta)$ parameterised by $\{ p(X_i \cbar Z_i) \} \cup \{p(Z_i \cbar Z_{\pa(i)})\}$, and $N_i$ quadrature points $\intz^{(i)}$ with weights $\intw^{(i)}$ for each $Z_i$ \\
\textbf{Complexity} $\gO(N^2)$, where $N = \max N_i$
\end{flushleft}
\begin{algorithmic}[1]
\State $S \gets \textsc{EmptyStackOfRegions}()$
\For{$c$ \textbf{in} $\textsc{PostOrderTraversal}(\gG)$} 
\If{$c$ \textbf{is} $\inputunit(p(X_i \cbar Z_i))$} 
    \State $R \gets \big[ \inputunit \, \big(p(X_i \cbar \intz[j]^{(i)})\big) \big]_{j=1}^{N_i}$
\ElsIf{$c$ \textbf{is} $\intu$}
    \State $R \gets\textsc{pop}(S)$ 
    \State $R \gets \big[\!\bigoplus \big(R, \big[ \intw[k]^{(i)}\,p(\intz[k]^{(i)} \cbar \intz[j]^{(\pa(i))}) \big]_{k=1}^{N_i} \big)\big]_{j=1}^{N_{\pa(i)}}$
\Else
    \State $Q \gets$ pop $\big| \inscope(c) \big|$ regions from $S$
    \State $R \gets \big[\!\bigotimes \big(\big[Q[i][j]\big]_{i=1}^{\textsc{len}(Q)}\big)\big]_{j=1}^{\textsc{len}(Q[0])}$
\EndIf
\State $\textsc{push}(S, R)$
\EndFor
\State \Return $\textsc{pop}(S)[0]$ \Comment{\QPC root unit}
\end{algorithmic}
\end{algorithm}
\end{minipage}\hfill
\begin{minipage}[b]{.48\textwidth}
  \includegraphics[scale=0.44]{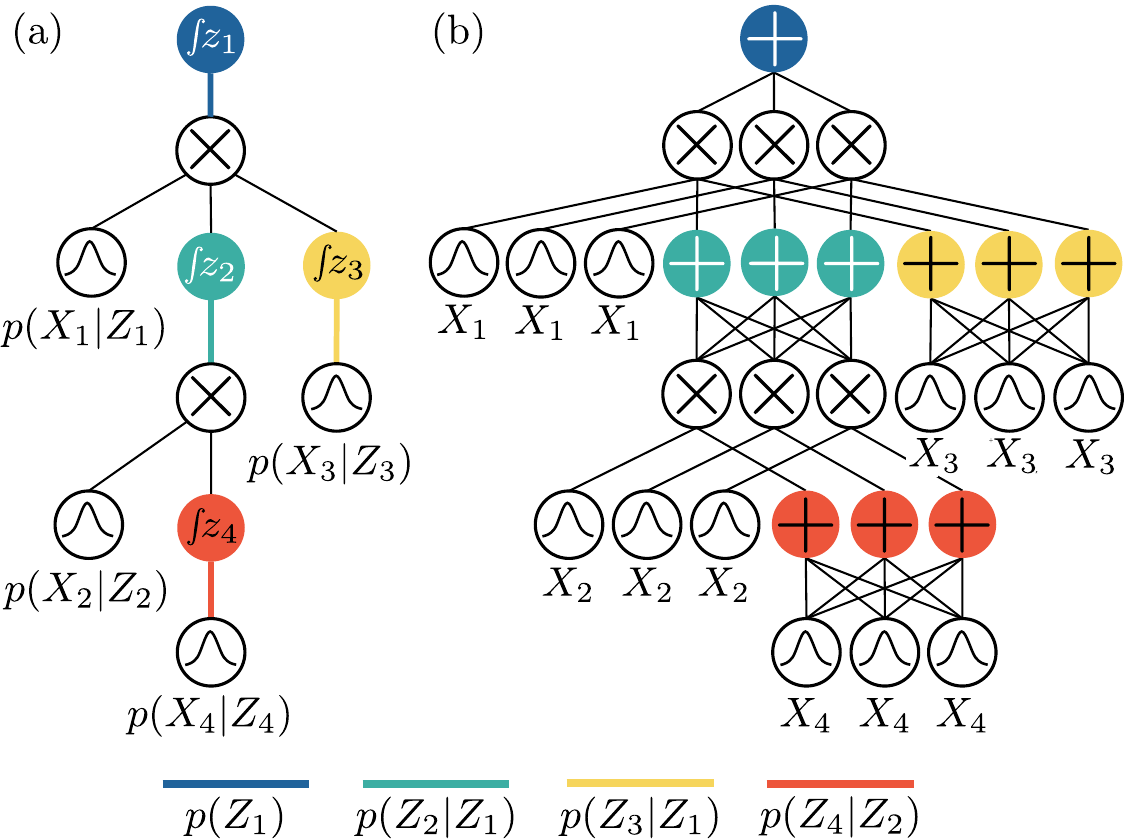}
  \captionof{figure}{
  \textbf{\PICs $\rightarrow$ \QPCs} A tree-shaped PIC in (a) and one of its possible \QPC obtained via \cref{alg:pic2qpc} using $N_i=3$ in (b).
  Following color-coding, every integral unit in (a)~has its incoming edge labelled with one of the conditional densities reported at the bottom.
  Both circuits are smooth and structured-decomposable.
  }
  \label{fig:pic}
\end{minipage}
\end{figure*}

\paragraph{Integral Circuits (\ICs)}
An \emph{integral circuit} $\gC \equiv (\gG, \Theta)$ is a \emph{symbolic} computational DAG $\gG$ parameterised by $\Theta$ that represents a function $\gC(\rmX) = \int \gC(\X, \rvz) \mathrm{d}\rvz$ over input variables $\rmX = \{X_i\}$ and auxiliary \emph{continuous} variables $\Z = \{Z_i\}$, where $\Z \cap \X = \varnothing$.
\ICs have four unit types: input \inputunit, integral \circled{$\smallint$}, sum $\bigoplus$, and product $\bigotimes$ units.
Specifically:
\begin{itemize}[leftmargin=*,itemsep=1pt,topsep=0pt]

    \item An input unit $c$ represents a function $f_c(\X_c, \Z_c) \rightarrow \sR$, where $\X_c \subseteq \X$ and $\Z_c \subseteq \Z$, and receives as input an observation $\x_{c}^{\prime}$ for variables $\X_{c}^{\prime} \subseteq \X_c$.
    The unit outputs the function $f_c[\x_{c}^{\prime} \setminus \X_{c}^{\prime}]$, where $[v \setminus V]$ is a syntactic substitution.\footnote{For example, applying the substitution $[2 \setminus A]$ to $f(A,B,C)=A^2+B-3C$ delivers $f(B,C)=2^2+B-3C$.}

    \item An integral unit $c$ has a function $f_c(\widehat{\Z}_c,\widecheck{\Z}_c) \rightarrow \sR$ attached, where $(\widehat{\Z}_c,\widecheck{\Z}_c) \subseteq \Z$ and $\widecheck{\Z}_c$ are its integration variables.
    The unit receives a function $f(\X_d,\widehat{\Z}_d,\widecheck{\Z}_d)$ from its only input unit $d$, where $\X_d\subseteq\X$, $(\widehat{\Z}_d,\widecheck{\Z}_d) \subseteq \Z$, and $\widecheck{\Z}_d = \widecheck{\Z}_c$, and outputs the function $\int f_c(\widehat{\Z}_c,\widecheck{\rvz}_c) \, f(\X_d,\widehat{\Z}_d,\widecheck{\rvz}_d) \, \mathrm{d}\widecheck{\rvz}_c$.

    \item A sum (resp.\ product) unit $c$ outputs a weighted sum (resp.\ product) of its incoming functions, and therefore a function in turn.
\end{itemize}

\IC parameters $\Theta$ comprise sum unit parameters, and input and integral unit functions $f_c$.
The scope of an input unit $c$ is $\X_c$, and the scope of an inner unit is the union of the scopes of its input units.
As such, the structural properties apply to \ICs as they do to PCs.

\paragraph{Probabilistic Integral Circuits (\PICs)}
A \PIC is a smooth and decomposable \IC whose input and integral unit functions $f_c$ are (conditional) probability density functions.
As such, \PIC auxiliary variables $\Z$ are continuous LVs of the hierarchical mixture the circuit represents.

We will assume that each integral unit has attached (i)~a single \emph{distinct} LV $Z_i$ as variable of integration, and (ii)~a (conditional) density of the form $p(Z_i \cbar \Z^{\prime})$, where $\rmZ^{\prime} \subseteq \rmZ \setminus Z_i$.
Using univariate LVs as integration variables allows for straightforward application of standard quadrature rules.
We represent an integral unit having integration variable $Z_i$ as $\intu$, and we label its incoming edge with $p(Z_i \cbar \Z^{\prime})$. 
An example \PIC is shown in \autoref{fig:pic}(a), where input units define conditional probabilities $p(X_i \cbar Z_i)$, and each integral unit is parameterized by the conditional $p(Z_i \cbar Z_{\pa(i)})$.

A \PIC is therefore a \emph{symbolic} computational graph representing a multivariate integral defined by nested univariate integrations.
As such, inference involves iteratively integrating out one LV while treating others as constants, which is akin to variable elimination \citep{darwiche2009modeling}.\footnote{Note that it may be possible to integrate out more LVs in parallel, like $Z_2$ and $Z_3$ in \autoref{fig:pic}(a).}
\PICs can in fact describe the operations needed to perform inference with continuous BNs, without \emph{necessarily} specifying how to solve the integrals.
\cref{alg:bn2pic} shows a possible way to convert an arbitrary BN to a smooth and structured-decomposable \PIC via variable elimination, where the resulting \PIC is always tree-shaped, regardless the BN structure. 
For example, converting the LTM in \autoref{fig:lt} delivers the \PIC in \autoref{fig:pic}(a).
Such conversion is important, as we can use \PIC structures for parameter learning and approximate inference (cf.\ \cref{sec:learnpic}).
Note that, unlike BNs, the conditioning variables of the density attached to an integral unit \emph{cannot} be deduced from the \PIC structure.

\paragraph{Tractable \PICs}
Inference in \PICs can be tractable, and therefore proceed in an efficient \emph{feedforward} fashion, similarly to standard PCs.
However, smoothness and decomposability are not sufficient anymore for tractability as we also need analytical integration to \emph{pass through} integral units.
Example of tractable \PICs are those representing the computation of Linear Gaussian BNs, i.e.\ BNs where all RVs are Gaussians and relationships among them are all linear \citep{bishop2006pattern, koller2009probabilistic}.
The output function of such \PICs would then be a multivariate Gaussian, as we show with a step-by-step \PIC computation in \autoref{app:lgm}.

\subsection{\QPCs approximate intractable \PICs}
\label{sec:qpc}

When the computation in integral units cannot be done in analytical form, \PICs become intractable. 
We propose numerical integration as a solution for such intractable scenarios.
Specifically, the application of \textbf{Q}uadrature rules to approximate intractable \PICs, results in standard PCs which we call \textbf{Q}PCs.
This way, we can \emph{materialise} the symbolic computational graph of a \PIC with a \QPC that numerically approximates it.
Despite \QPCs being standard PCs, their LV interpretation is not categorical anymore but rather \emph{ordinal}, as the circuit represents the application of a quadrature rule on a \emph{continuous} intractable model.

Numerically approximating an intractable \PIC begins with the application of a quadrature rule over the outermost integral unit (e.g., the root unit for \PIC in \autoref{fig:pic}(a)) and then proceeds iteratively with the lower ones.
Specifically, for each integration point of an outer integral we would have many for the inner ones, and so on iteratively.
Therefore, the selection of integration points plays a crucial role in our numerical approximation.

On the one hand, if for every integration problem involving a specific LV $Z_i$ we \emph{always} use the same static integration points, we can materialise \PICs into \QPCs with a high re-use of computational units, since same integrand evaluations will be required multiple times.
This is the case for \QPCs resulting from the application of a \emph{static} quadrature rule.
This is detailed in \cref{alg:pic2qpc}, where we show how a specific type of tree-shaped \PIC can be converted into a \QPC compiled in a bottom-up fashion.
For instance, if we were to materialise the PIC in \autoref{fig:pic}(a) following \cref{alg:pic2qpc} and using $N=3$ static integration points for each integral unit, we would obtain the structure in \autoref{fig:pic}(b).
Note that applying \cref{alg:pic2qpc} on tree-shaped, smooth and (structured-)decomposable \PICs, results in non-tree \QPCs with the same structural properties.

On the other hand, if we do not maintain uniformity in our integration point selection, unit re-usage is not possible, and we may need an exponentially big \QPC to represent the computation of the resulting numerical approximation.
For instance, Gauss-Hermite quadrature chooses the integration points w.r.t.\ characteristics of the integrand itself, therefore prohibiting re-usage of computational units.
Even more dramatically, in cases where the selection of integration points is adaptive, a dependency emerges with the data point $\rvx$.
That is because adaptive quadratures can iteratively require more integration points in certain sub-intervals where the density for $\rmX$ is relatively difficult to approximate.
Consequently, this prevents us from materializing a circuit that can be arbitrarily good for any $\rvx$.

\paragraph{\PICs \& LTMs}
Numerically approximating \PICs is particularly effective when the conditional distributions attached to integral units are not conditioned on a large number of LVs.
That is especially the case for \PICs representing inference of continuous LTMs. 
Therefore, all the latent conditional distributions of such \PIC type are only conditioned on a single LV, except for the root integral unit, which defines a prior.
We show how LTMs (cf. \autoref{fig:lt}) can be converted to PICs (cf. \autoref{fig:pic}(a)) in \cref{alg:bn2pic}.
Furthermore, \cref{alg:quad} shows how we can generally apply nested univariate quadratures to a tree-shaped \PIC.

\subsection{Learning neural \PICs via quadrature}
\label{sec:learnpic}

We here present a methodology to (i)~parameterise \PICs using light-weight neural energy-based models and (ii)~learn the resulting model via numerical integration.
It can be applied to every \PIC structure, but is particularly suitable for those where LV interactions follow a common pattern.
Specifically, we will describe and show its application to \PICs compiled from LTMs, since \emph{all} their LVs are in a pair-wise relationship and therefore only involve univariate integrations.
%
%
So we are interested in approximating integrals of the form $\smallint_{a}^b q(x_i,z_i,z_{\pa(i)}) dz_i$, with $q(x,\widecheck{z},\widehat{z})\equiv p(\widecheck{z} \cbar \widehat{z})p(x \cbar \widecheck{z})$, such that the integration error
\begin{equation*}\textstyle
    \varepsilon_N(x, \widehat{z}) = \left\lvert\int_{a}^{b} q(x,\widecheck{z},\widehat{z}) \mathrm{d}\widecheck{z} - \sum_{i=1}^N \intw[n] q(x,\intz[n],\widehat{z}) \right\lvert,
\end{equation*}
goes to zero as $N \rightarrow \infty$.

\paragraph{Parameterising integral units}
Every latent distribution $p(Z_i \cbar Z_{\pa(i)})$ attached to integral unit $\intu$ is parameterised by a light-weight energy-based model (EBM) \citep{lecun2006tutorial, song2020score} $\sumnet$ outputting an energy value for a \emph{conditional} input state $(z_i, z_{\pa(i)})$ of its continuous LVs  (cf.\ \autoref{fig:quad}(c)).
To promote the application of static quadrature rules, we will assume that every LV has support $[-1, 1]$, such that pairs $(z_i, z_{\pa(i)})$ belong to $[-1, 1]^2$.
Every neural net $\sumnet$ parametrizes a conditional univariate 
energy-based model:
\begin{align}
p(z_i \cbar z_{\pa(i)}) = 
\frac{\exp(-\sumnet(z_i, z_{\pa(i)}))}{\int_{-1}^{+1} \exp(-\sumnet(z, z_{\pa(i)}))dz},
\label{eq:ebm}
\end{align}
where $\sumnet: [-1, 1]^2 \rightarrow \sR_{\geq 0}$ and the denominator is the normalisation constant for $p(Z_i \cbar z_{\pa(i)})$.
We constrain the output of $\sumnet$ to be non-negative using a softplus non-linearity, so that the unnormalized density, i.e.\ the numerator, is bounded by one, as that improves training stability \citep{nash2019autoregressive}.

\paragraph{Parameterising input units}
We use a similar modelling for input units.
Every conditional input distribution $p(X_i \cbar z_i)$ is parameterised by a neural net $\innet$ taking as input a state for its parent, $z_i \in [-1, 1]$, and returning a valid parametrisation of a parametric univariate distribution (cf. \autoref{fig:quad}(d)), similarly to a small VAE decoder \citep{kingma2013auto}.
For instance, if $X_i$ were a Gaussian RV, then:
\begin{align}
\textstyle
p(X_i \cbar z) = \gN(X_i \cbar \mu_z, \sigma_z), 
\end{align}
where $(\mu_z, \sigma_z) = \innet(z)$.
In contrast to standard VAE and flow models, we can easily handle heterogeneous features---i.e.\ both discrete and continuous---by simply designing a specific neural net $\innet$ to output a valid parameterisation for the desired $X_i$ input distribution.

\paragraph{Fourier Features}
Fourier Feature Layers (FFLs) are an important ingredient for the neural nets $\sumnets$ and~$\innets$.
FFLs \citep{tancik2020fourier} enable  Multi-Layer Perceptrons (MLPs) to learn high-frequency functions in low-dimensional problem domains and are usually used as first layers of coordinate-based MLPs as in \autoref{fig:quad}(c-d).
FFLs transform input $\rvx \in \sR^n$ to 
\[
[ \cos (2 \pi \rvf_1^{\intercal} \rvx),  \sin (2 \pi \rvf_1^{\intercal} \rvx), \dots, \cos (2 \pi \rvf_k^{\intercal} \rvx),  \sin (2 \pi \rvf_k^{\intercal} \rvx)],
\] where $k$ is a hyperparameter and vectors $\rvf_i \in \sR^n$ are non-learnable, randomly initialized parameters.

FFLs have two main benefits: (i) They allow learning more expressive distributions by avoiding over-smoothing,  and (ii) they significantly reduce the total count of optimizable parameters when used instead of conventional linear layers as the initial layers in MLPs.
We show the effect of using FFLs in \cref{sec:fourier_vs_no_fourier}.

\paragraph{Learning via numerical integration}
We train neural \PICs via \emph{static} numerical integration.
Specifically, we optimise neural net parameters $\phi$ by materialising neural \PICs into \QPCs via \cref{alg:pic2qpc} and then minimizing the negative log-likelihood under the materialised model at each training step.
However, the algorithm cannot be directly applied, as the densities attached to integral units are EBMs $\sumnets$, for which we need the normalisation constants.
Nevertheless, by our design, they can be approximated arbitrarily well via quadrature due to the univariate nature of the density in \autoref{eq:ebm}.
Therefore, there are two distinct approximations involved, (i)~one concerning the normalisation constants for $\sumnets$, and (ii)~another for the integral units.
While these approximations can be approached independently, it is computationally cheaper if we use the same quadrature rule for both, especially during training, as the required function evaluations would then be the same for both.
Therefore, this materialisation-by-approximation of \QPCs is central for learning \PICs, and below we describe how this can be done \emph{efficiently}.

Since the support of $\sumnet$ and $\innet$ is $[-1, 1]$, we use the same \emph{static} integration points $\intz \in [-1, 1]^{N}$ and weights $\intw  \in \sR^{N}$ for every integral unit.
Leveraging GPU acceleration, we start by generating the tensor of energy values $\gE \in \sR^{D \times N \times N}$ \emph{in one shot} as
\begin{equation}
\textstyle
\gE_{ijk} = \sumnet(\intz[k], \intz[j]),
\label{eq:energy-tensor}
\end{equation}
where $i \in [D]$ and $j,k \in [N]$.
Then, we use quadrature to approximate the normalisation constants as
\begin{align}
\textstyle
\gN_{ij} = \sum_{k=1}^N \intw[k] \exp(-\gE_{ijk}),
\label{eq:normconst}
\end{align}
which we then use to move from $\gE$ to a tensor of approximated probability density values $\gP$ as
\begin{align}
\textstyle
\gP_{ijk} = \frac{\exp(-\gE_{ijk})}{\gN_{ij}} \approx p(Z_i = \intz[k] \cbar Z_{\pa(i)} = \intz[j]).
\label{eq:probability-tensor}
\end{align}
Note that sums over the third dimension of $\gP$ do \emph{not} return 1.
Finally, we approximate the integral units w.r.t.\ the same quadrature rule used in \autoref{eq:normconst}.
Specifically, we move from $\gP$ to a tensor of sum unit parameters as
\begin{equation}
\textstyle
\gS_{ijk} = \intw[k] \gP_{ijk},
\end{equation}
where $k \in [N]$.
Note that sums over the third dimension of $\gS$ now return $1$.
We perform such computation in the log space to ensure numerical stability.
Concisely, we denote the materialisation of \emph{all} sum unit parameters $\gS$ w.r.t.\ integration points $\intz$ and weights $\intw$ as $\sumnets(\intz, \intw)$.
Similarly, we materialise \emph{all} input unit parameters in a tensor $\gI \in \sR^{D \times N \times I}$ as
\begin{equation}
\textstyle
\gI_{ij} = \innet(\intz[j]) \in \sR^I,
\end{equation}
where $I$ is the number of parameters per input unit.\footnote{We assume, for ease of presentation, that \emph{all} input units of a circuit have the same parametric distribution.}
Concisely, we denote the materialisation of input unit parameters w.r.t. integration points $\intz$ as $\innets(\intz)$.
The training flow is sketched in \autoref{fig:quad}(e).

Under a static quadrature regime, both $\sumnet$ and $\innet$ can be run in parallel for every $i$, delivering a fast materialisation of \QPCs.
Importantly, the circuit can be materialized for any $N$, delivering arbitrarily wide \QPCs.
Furthermore, the neural nets $(\sumnets, \innets)$ represent a very compact representation for the materialised \QPC when a large $N$ is used.
That is because the size of $\sumnets$ and $\innets$ is independent from the $N$ we choose.
Moreover, similar to \cite{shih2021hyperspns}, not only can we be memory-efficient during storage, but we can also be memory-efficient during evaluation.
In fact, we materialize \QPC regions on the fly, and erase them from memory when they become unnecessary.
We conjecture that training via numerical integration sufficiently regularises the neural nets $\sumnets$ and $\innets$ so that the overall model remains amenable to numerical integration.

\section{Experiments}
\label{sec:exp}

\begin{figure}
  \begin{minipage}[c]{0.14\textwidth}
    \includegraphics[scale=0.45]{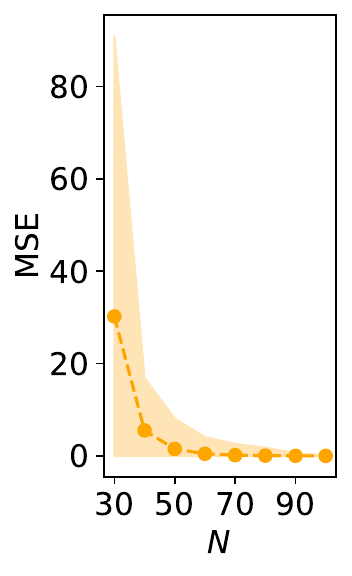}
  \end{minipage}\hfill
  \begin{minipage}[c]{0.30\textwidth}
    \caption{
    \textbf{Increasing the number of integration points reduces the approximation error.} MSE (y-axis) between linear Gaussian LTM log-likelihoods and \QPC approximations at increasing number of integration points $N$ (x-axis) averaged over 50 different LTMs with 256 nodes.}
    \label{fig:qpc_gauss}
  \end{minipage}
  \vspace{-5mm}
\end{figure}

\subsection{A tractable sanity check}
\label{sec:sanitycheck}

We start off our experiments showing empirically that \QPCs can be used to approximate (tractable) continuous LV models.
We do this by using linear Gaussian LTMs as tractable ground-truth models and show, as a sanity check, that these can be approximated arbitrarily well with \QPCs.
Specifically, we convert these LTMs to \PICs using \cref{alg:bn2pic} and apply \cref{alg:pic2qpc} to approximate them by \QPCs.
To measure the goodness of our approximation, we generate 1,000 samples from the ground-truth model and then use the MSE between their true log-likelihoods and their \QPC approximations.
We provide more detail in \autoref{app:sanitycheck}.
Results are sketched in \autoref{fig:qpc_gauss}.

\subsection{Distribution estimation}

In the following experiments, we use the latent tree structure provided by the Hidden Chow-Liu Tree algorithm \citep{liu2021tractable}.
These structures are heuristically derived by introducing LVs in classical Chow-Liu Trees (CLTs).
Specifically, the structure is built in three steps: (i) first, a tree over input variables $\rmX = \{ X_i \}_{i=1}^{D}$ is learned via the CLT algorithm \citep{chow1968approximating}, (ii) then, every node $X_i$ in the tree is replaced with LV $Z_i$, and (iii) finally, every node $X_i$ is reintroduced in the tree as child of $Z_i$.
Assuming LVs to be categorical, these LTMs can be compiled into smooth and structured-decomposable PCs---precisely resembling the one in \autoref{fig:pic}(b)---which are called HCLTs.
They demonstrated competitive performance across many different tasks in recent literature \citep{liu2022lossless, liu2021tractable, liu2023scaling, dang2023tractable, liu2023understanding}.
Our code is available at  \href{https://github.com/gengala/pic}{github.com/gengala/pic}.

\subsubsection{MNIST Family}
Following \citet{DangNeurIPS22}, we evaluate \PICs on the MNIST-family image datasets which includes MNIST \citep{mnist}, EMNIST (and its 4 splits) \citep{cohen2017emnist}, and FashionMNIST \citep{xiao2017fashion}.
We use the average test-set bits-per-dimension $\mathsf{bdp}(\gC(\rvx)) = -\log_2(\gC(\rvx)) / D$ as the evaluation metric.

\paragraph{Setting}
Our main reference is the HCLT model \citep{liu2021tractable} as applying a static quadrature on \PICs results in \QPCs having \emph{exactly} the same structure.
We ablate using $N=\{16, 32, 64, 128\}$ and batch size $B=\{64, 128, 256\}$.
The hyperparameter $N$ refers to the number of integration points when associated to \QPCs and to the number of latent categorical states when associated to HCLTs.
We train HCLTs using stochastic mini-batch Expectation-Maximization.
Both the EM step size (used for HCLTs) and learning rate (used for \PICs) are annealed using the Cosine Annealing with Warm Restarts scheduler \citep{loshchilov2016sgdr} to avoid sensitivity to initialization which often leads to convergence issues in PCs.
For a fair comparison, we re-implement HCLTs in pytorch and achieve \emph{better results} than those reported by \cite{liu2022lossless} and \cite{DangNeurIPS22}.
We report more training details and experiments in \autoref{app:exp}.

\begin{figure*}[!ht]
	\hfill
	{\footnotesize
		\begin{tabular}{lcc|cccccc}
			\toprule
			& \QPC & HCLT & Sp-PC & RAT & IDF & BitS & BBans & McB \\
			\midrule
			\textsc{mnist}      & \textbf{1.18} & 1.21 & \underline{1.14} & 1.67 & 1.90 & 1.27 & 1.39 & 1.98 \\
			\textsc{f-mnist}    & \underline{\textbf{3.27}} & 3.34 & \underline{3.27}  & 4.29 & 3.47 & 3.28 & 3.66 & 3.72 \\
			\textsc{emn-mn}     & \textbf{1.66} & 1.70 & \underline{1.52}  & 2.56 & 2.07 & 1.88 & 2.04 & 2.19 \\
			\textsc{emn-le}     & \textbf{1.70} & 1.75 & \underline{1.58}  & 2.73 & 1.95 & 1.84 & 2.26 & 3.12 \\
			\textsc{emn-ba}     & \textbf{1.73} & 1.78 & \underline{1.60}  & 2.78 & 2.15 & 1.96 & 2.23 & 2.88 \\
			\textsc{emn-by}     & \textbf{1.67} & 1.73 & \underline{1.54}  & 2.72 & 1.98 & 1.87 & 2.23 & 3.14 \\
			\bottomrule
		\end{tabular}
	}
	\hfill
	\includegraphics[align=c,scale=0.34]{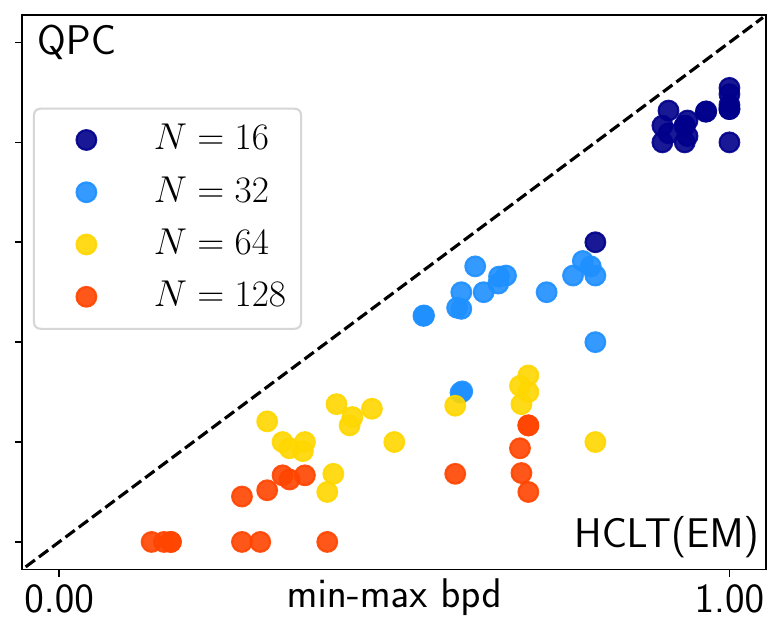}
	\hfill\null
	\caption{
		\textbf{\QPCs systematically outperform PCs trained via EM or SGD.}
		Table (left): Best average test-set bpd for the \textsc{mnist}-famility datasets.
		We compare against HCLT \citep{liu2021tractable}, SparsePC \citep{DangNeurIPS22} RAT-SPN \citep{peharz2020random}, IDF \citep{hoogeboom2019integer}, BitSwap \citep{kingma2019bit}, BBans \citep{townsend2019practical} and McBits \citep{ruan2021improving}.
  		\QPC results are in bold if better than HCLTs, whereas global best results are underlined.
		\QPC and HCLT results are averaged over 5 different runs; the other results are taken from \citet{DangNeurIPS22}.
		Scatter plot (right): bpd results for \QPCs (y-axis) and HCLTs (x-axis) paired by $B$-$N$ hyperparameter configuration and (min-max) normalized for every MNIST-family dataset.
        A similar trend occurs for binomial input units (cf.\ \autoref{app:exp}).
	}
	\label{fig:mnist}
\end{figure*}

\paragraph{Results}
\autoref{fig:mnist} reports the best average test-set bpd for the MNIST-famility.
Remarkably, we outperform \emph{all} the baselines except SparsePC \citep{DangNeurIPS22}, which, however, uses a more sophisticated structure learner, which iteratively prunes and grows the computational graph of an HCLT.

Basing our comparison on standard HCLTs, our main reference, not only \QPCs outperform them in terms of best bpd, but they do that consistently for every $B$-$N$ configuration as we show in the scatter plot in \autoref{fig:mnist}.
The same trend occurs when comparing \QPCs with HCLTs trained with Adam \citep{kingma2014adam} and using binomial input units (cf.\ \autoref{app:exp}).
\QPCs also outperform shallow CMs of PCs \citep{correia2023continuous} as they only score 1.28 (resp.\ 3.55) on \textsc{Mnist} (resp.\ \textsc{Fashion-Mnist}).
Note that when using $N=128$, \PICs have roughly half the parameters of their materialised \QPCs. 

\subsubsection{Penn Tree Bank}
Again following \citet{DangNeurIPS22}, we use the Penn Tree Bank dataset with standard processing from \citet{mikolov2012subword}, which contains around 5M characters and a character-level vocabulary size of 50.
The data is split into sentences with a maximum sequence length of 288.
We compare with four competitive baselines: SparsePC \citep{DangNeurIPS22}, Bipartite flow \citep{tran2019discrete} and latent flows \citep{ziegler2019latent} including AF/SCF and IAF/SCF, since they are the only comparable work with non-autoregressive language modeling.
\autoref{tab:ptb} show that \QPCs remarkably outperform all baselines by a considerable margin.

\begin{table}
\centering
\caption{
Character-level language modeling results on Penn Tree Bank in test-set bpd.
Competitor results are taken from \citet{DangNeurIPS22}}
\label{tab:ptb}
\begin{tabular}{cc|ccc}
\hline
\QPC & Sp-PC & B-flow & AF/SCF  & IAF/SCF \\
\hline
\textbf{1.22}  & 1.35  & 1.38  & 1.46  &  1.63 \\   
\hline
\end{tabular}
\end{table}

\section{Discussion \& Conclusion}

\paragraph{Related Work}
Like HyperSPN \citep{shih2021hyperspns} and ConditionalSPN \citep{shao2020conditional}, we materialise circuit parameters using neural nets.
However, different from HyperSPN, we can materialise an \emph{arbitrary big} \QPCs whose size is independent of the size of the neural nets $(\sumnets, \innets)$.
That is because HyperSPN still works under the categorical LV interpretation, and therefore is constrained by design to output a fixed number of circuit parameters.

Closer to ours, \citet{correia2023continuous} work under the continuous LV interpretation as we do, and propose a simple CM of PCs, where the parameters of each PC are dependent on a continuous multivariate LV via a learnable function parametrised by a neural net, which resembles a  VAE decoder.
These models---like \PICs---are inherently intractable, but for small latent dimensions, can be accurately approximated via numerical integration methods, which allows learning via regular backpropagation.
The numerical approximation eventually yields a discrete mixture of PCs, which is a PC in itself.
However, the application of CMs by \citet{correia2023continuous} is designed and limited for (large) root units only, and their usage as internal units is not explored.

\vspace{-2mm}
\paragraph{Model interpretation}
One can interpret our model in two distinct ways \citep[cf.][]{correia2023continuous}.
The first is to interpret it as a \emph{factory method}, whereby each finite set of LV values (integration points) yields a PC.
The second is to take it as an intractable continuous LV model, but one that is amenable to numerical integration.
At test time, we are free to change the integration method and the number of integration points if more or less precision is needed.
%

\paragraph{Broader Impact}
The use of continuous LVs in \PICs can unlock new learning schemes for PCs, such as variational inference, adversarial and self-supervised learning.
These learning schemes are not applicable to PCs directly, as they do \emph{not} provide differentiable sampling.
This is because their backward pass is not differentiable, as sum units involve categorical sampling.
Recently, \citet{lang2022elevating} used the Gumbel-Softmax trick \citep{jang2016categorical}, combined with a smooth children-interpolation to bypass the issue, but with very little practical success.
Instead, \PICs are more directly compatible with differentiable sampling and therefore amenable to the previously mentioned learning schemes.
That is possible if the neural nets $\sumnet$ allow for differentiable sampling.
For this work, the neural nets $\sumnets$ are EBMs, which are incapable of differentiable sampling.
However, for instance, replacing EBMs with Normalizing Flows would already be sufficient to allow for differentiable sampling in \PICs.

\paragraph{Limitations \& Future Work}
Training \PICs requires a little overhead both in space and time (cf.\ \autoref{app:exp}).
For future work, we plan to investigate higher latent dimensionalities, arbitrary \PIC structures, differentiable sampling and pruning methods.
This work shows new means to perform inference with continuous LVs with a possibly complex association structure.
It is left open the possibility to learn the better structures, beyond HCLTs.
This work may also open research on inference algorithms for models such as continuous Bayesian nets and to extend models such as variational auto-encoders to represent latent structures that are amenable to marginalization via quadrature rules.

\section*{Acknowledgments}
The Eindhoven University of Technology authors thank the support from the Eindhoven Artificial Intelligence Systems Institute and the Department of Mathematics and Computer Science of TU Eindhoven. 
Robert Peharz's research was supported by the Graz Center for Machine Learning (GraML).
Cassio de Campos thanks the support of EU European Defence Fund Project KOIOS (EDF-2021-DIGIT-R-FL-KOIOS) and Dutch NWO Perspectief 2022 Project PersOn (P21-03).
Antonio Vergari was supported by the “UNREAL” project (EP/Y023838/1) selected by the ERC and funded by UKRI EPSRC.

\bibliographystyle{plainnat}
\bibliography{main}

\begin{thebibliography}{54}
\providecommand{\natexlab}[1]{#1}
\providecommand{\url}[1]{\texttt{#1}}
\expandafter\ifx\csname urlstyle\endcsname\relax
  \providecommand{\doi}[1]{doi: #1}\else
  \providecommand{\doi}{doi: \begingroup \urlstyle{rm}\Url}\fi

\bibitem[Abramowitz et~al.(1988)Abramowitz, Stegun, and Romer]{abramowitz1988handbook}
Milton Abramowitz, Irene~A Stegun, and Robert~H Romer.
\newblock Handbook of mathematical functions with formulas, graphs, and mathematical tables, 1988.

\bibitem[Bender et~al.(2023)Bender, Shi, Niethammer, and Oliva]{bender23continuously}
Christopher~M Bender, Yifeng Shi, Marc Niethammer, and Junier Oliva.
\newblock Continuously parameterized mixture models.
\newblock In \emph{Proceedings of the 40th International Conference on Machine Learning}, volume 202 of \emph{Proceedings of Machine Learning Research}, pages 2050--2062. PMLR, 23--29 Jul 2023.
\newblock URL \url{https://proceedings.mlr.press/v202/bender23a.html}.

\bibitem[Bishop and Nasrabadi(2006)]{bishop2006pattern}
Christopher~M Bishop and Nasser~M Nasrabadi.
\newblock \emph{Pattern recognition and machine learning}, volume~4.
\newblock Springer, 2006.

\bibitem[Bond-Taylor et~al.(2021)Bond-Taylor, Leach, Long, and Willcocks]{bond2021deep}
Sam Bond-Taylor, Adam Leach, Yang Long, and Chris~G Willcocks.
\newblock Deep generative modelling: A comparative review of vaes, gans, normalizing flows, energy-based and autoregressive models.
\newblock \emph{IEEE transactions on pattern analysis and machine intelligence}, 2021.

\bibitem[Butz et~al.(2020)Butz, S.~Oliveira, and Peharz]{pmlr-v138-butz20a}
Cory Butz, Jhonatan S.~Oliveira, and Robert Peharz.
\newblock Sum-product network decompilation.
\newblock In Manfred Jaeger and Thomas~Dyhre Nielsen, editors, \emph{Proceedings of the 10th International Conference on Probabilistic Graphical Models}, volume 138 of \emph{Proceedings of Machine Learning Research}, pages 53--64. PMLR, 23--25 Sep 2020.
\newblock URL \url{https://proceedings.mlr.press/v138/butz20a.html}.

\bibitem[Choi et~al.(2011)Choi, Tan, Anandkumar, and Willsky]{choi2011learning}
Myung~Jin Choi, Vincent~YF Tan, Animashree Anandkumar, and Alan~S Willsky.
\newblock Learning latent tree graphical models.
\newblock \emph{Journal of Machine Learning Research}, 12:\penalty0 1771--1812, 2011.

\bibitem[Choi et~al.(2020)Choi, Vergari, and Van~den Broeck]{choiprobabilistic}
YooJung Choi, Antonio Vergari, and Guy Van~den Broeck.
\newblock Probabilistic circuits: A unifying framework for tractable probabilistic models.
\newblock \emph{preprint}, 2020.

\bibitem[Chow and Liu(1968)]{chow1968approximating}
CKCN Chow and Cong Liu.
\newblock Approximating discrete probability distributions with dependence trees.
\newblock \emph{IEEE transactions on Information Theory}, 14\penalty0 (3):\penalty0 462--467, 1968.

\bibitem[Cohen et~al.(2017)Cohen, Afshar, Tapson, and Van~Schaik]{cohen2017emnist}
Gregory Cohen, Saeed Afshar, Jonathan Tapson, and Andre Van~Schaik.
\newblock Emnist: Extending mnist to handwritten letters.
\newblock In \emph{2017 international joint conference on neural networks (IJCNN)}, pages 2921--2926. IEEE, 2017.

\bibitem[Correia et~al.(2023)Correia, Gala, Quaeghebeur, de~Campos, and Peharz]{correia2023continuous}
Alvaro~HC Correia, Gennaro Gala, Erik Quaeghebeur, Cassio de~Campos, and Robert Peharz.
\newblock Continuous mixtures of tractable probabilistic models.
\newblock In \emph{Proceedings of the AAAI Conference on Artificial Intelligence}, volume~37, pages 7244--7252, 2023.

\bibitem[Dang et~al.(2022)Dang, Liu, and Van~den Broeck]{DangNeurIPS22}
Meihua Dang, Anji Liu, and Guy Van~den Broeck.
\newblock Sparse probabilistic circuits via pruning and growing.
\newblock In \emph{Advances in Neural Information Processing Systems 35 (NeurIPS)}, dec 2022.

\bibitem[Dang et~al.(2023)Dang, Liu, Wei, Sankararaman, and Van~den Broeck]{dang2023tractable}
Meihua Dang, Anji Liu, Xinzhu Wei, Sriram Sankararaman, and Guy Van~den Broeck.
\newblock Tractable and expressive generative models of genetic variation data.
\newblock \emph{bioRxiv}, pages 2023--05, 2023.

\bibitem[Darwiche(2009)]{darwiche2009modeling}
Adnan Darwiche.
\newblock \emph{Modeling and reasoning with Bayesian networks}.
\newblock Cambridge university press, 2009.

\bibitem[Davis and Rabinowitz(2007)]{davis2007methods}
Philip~J Davis and Philip Rabinowitz.
\newblock \emph{Methods of numerical integration}.
\newblock Courier Corporation, 2007.

\bibitem[{de Campos} et~al.(2020){de Campos}, Stamoulis, and Weyland]{DECAMPOS2020104627}
Cassio~P. {de Campos}, Georgios Stamoulis, and Dennis Weyland.
\newblock A structured view on weighted counting with relations to counting, quantum computation and applications.
\newblock \emph{Information and Computation}, 275:\penalty0 104627, 2020.
\newblock \doi{10.1016/j.ic.2020.104627}.

\bibitem[Ghahramani(2015)]{Ghahramani2015}
Zoubin Ghahramani.
\newblock Probabilistic machine learning and artificial intelligence.
\newblock \emph{Nature}, 521\penalty0 (7553):\penalty0 452--459, 2015.

\bibitem[Goodfellow et~al.(2014)Goodfellow, Pouget-Abadie, Mirza, Xu, Warde-Farley, Ozair, Courville, and Bengio]{goodfellow2014generative}
Ian~J Goodfellow, Jean Pouget-Abadie, Mehdi Mirza, Bing Xu, David Warde-Farley, Sherjil Ozair, Aaron~C Courville, and Yoshua Bengio.
\newblock Generative adversarial nets.
\newblock In \emph{NIPS}, 2014.

\bibitem[Hoogeboom et~al.(2019)Hoogeboom, Peters, Van Den~Berg, and Welling]{hoogeboom2019integer}
Emiel Hoogeboom, Jorn Peters, Rianne Van Den~Berg, and Max Welling.
\newblock Integer discrete flows and lossless compression.
\newblock \emph{Advances in Neural Information Processing Systems}, 32, 2019.

\bibitem[Jang et~al.(2016)Jang, Gu, and Poole]{jang2016categorical}
Eric Jang, Shixiang Gu, and Ben Poole.
\newblock Categorical reparameterization with gumbel-softmax.
\newblock \emph{arXiv preprint arXiv:1611.01144}, 2016.

\bibitem[Jaynes(2003)]{Jaynes2003}
E.~T. Jaynes.
\newblock \emph{{Probability Theory: The Logic of Science}}.
\newblock Cambridge University Press, 2003.

\bibitem[Kingma and Welling(2014)]{kingma2013auto}
D.~P. Kingma and M.~Welling.
\newblock Auto-encoding variational {Bayes}.
\newblock In \emph{International Conference on Learning Representations (ICLR)}, 2014.
\newblock arXiv:1312.6114.

\bibitem[Kingma and Ba(2014)]{kingma2014adam}
Diederik~P Kingma and Jimmy Ba.
\newblock Adam: A method for stochastic optimization.
\newblock \emph{arXiv preprint arXiv:1412.6980}, 2014.

\bibitem[Kingma et~al.(2019)Kingma, Abbeel, and Ho]{kingma2019bit}
Friso Kingma, Pieter Abbeel, and Jonathan Ho.
\newblock Bit-swap: Recursive bits-back coding for lossless compression with hierarchical latent variables.
\newblock In \emph{International Conference on Machine Learning}, pages 3408--3417. PMLR, 2019.

\bibitem[Koller and Friedman(2009)]{koller2009probabilistic}
Daphne Koller and Nir Friedman.
\newblock \emph{Probabilistic graphical models: principles and techniques}.
\newblock MIT press, 2009.

\bibitem[Lang et~al.(2022)Lang, Mundt, Ventola, Peharz, and Kersting]{lang2022elevating}
Steven Lang, Martin Mundt, Fabrizio Ventola, Robert Peharz, and Kristian Kersting.
\newblock Elevating perceptual sample quality in pcs through differentiable sampling.
\newblock In \emph{NeurIPS 2021 Workshop on Pre-registration in Machine Learning}, pages 1--25. PMLR, 2022.

\bibitem[LeCun et~al.(2006)LeCun, Chopra, Hadsell, Ranzato, and Huang]{lecun2006tutorial}
Yann LeCun, Sumit Chopra, Raia Hadsell, M~Ranzato, and Fujie Huang.
\newblock A tutorial on energy-based learning.
\newblock \emph{Predicting structured data}, 1\penalty0 (0), 2006.

\bibitem[Liu and Van~den Broeck(2021)]{liu2021tractable}
Anji Liu and Guy Van~den Broeck.
\newblock Tractable regularization of probabilistic circuits.
\newblock \emph{Advances in Neural Information Processing Systems}, 34, 2021.

\bibitem[Liu et~al.(2022)Liu, Mandt, and den Broeck]{liu2022lossless}
Anji Liu, Stephan Mandt, and Guy~Van den Broeck.
\newblock Lossless compression with probabilistic circuits.
\newblock In \emph{International Conference on Learning Representations}, 2022.
\newblock URL \url{https://openreview.net/forum?id=X_hByk2-5je}.

\bibitem[Liu et~al.(2023{\natexlab{a}})Liu, Zhang, and den Broeck]{liu2023scaling}
Anji Liu, Honghua Zhang, and Guy~Van den Broeck.
\newblock Scaling up probabilistic circuits by latent variable distillation.
\newblock In \emph{The Eleventh International Conference on Learning Representations}, 2023{\natexlab{a}}.
\newblock URL \url{https://openreview.net/forum?id=067CGykiZTS}.

\bibitem[Liu et~al.(2023{\natexlab{b}})Liu, Liu, Van~den Broeck, and Liang]{liu2023understanding}
Xuejie Liu, Anji Liu, Guy Van~den Broeck, and Yitao Liang.
\newblock Understanding the distillation process from deep generative models to tractable probabilistic circuits.
\newblock In \emph{International Conference on Machine Learning}, pages 21825--21838. PMLR, 2023{\natexlab{b}}.

\bibitem[Loshchilov and Hutter(2016)]{loshchilov2016sgdr}
Ilya Loshchilov and Frank Hutter.
\newblock Sgdr: Stochastic gradient descent with warm restarts.
\newblock \emph{arXiv preprint arXiv:1608.03983}, 2016.

\bibitem[Mari et~al.(2023)Mari, Vessio, and Vergari]{mari2023unifying}
Antonio Mari, Gennaro Vessio, and Antonio Vergari.
\newblock Unifying and understanding overparameterized circuit representations via low-rank tensor decompositions.
\newblock In \emph{The 6th Workshop on Tractable Probabilistic Modeling}, 2023.
\newblock URL \url{https://openreview.net/forum?id=1btutFdIya}.

\bibitem[Mikolov et~al.(2012)Mikolov, Sutskever, Deoras, Le, Kombrink, and Cernocky]{mikolov2012subword}
Tom{\'a}{\v{s}} Mikolov, Ilya Sutskever, Anoop Deoras, Hai-Son Le, Stefan Kombrink, and Jan Cernocky.
\newblock Subword language modeling with neural networks.
\newblock \emph{preprint (http://www. fit. vutbr. cz/imikolov/rnnlm/char. pdf)}, 8\penalty0 (67), 2012.

\bibitem[Nash and Durkan(2019)]{nash2019autoregressive}
Charlie Nash and Conor Durkan.
\newblock Autoregressive energy machines.
\newblock In \emph{International Conference on Machine Learning}, pages 1735--1744. PMLR, 2019.

\bibitem[Peharz(2015)]{Peharz2015a}
R.~Peharz.
\newblock \emph{Foundations of Sum-Product Networks for Probabilistic Modeling}.
\newblock PhD thesis, Graz University of Technology, 2015.

\bibitem[Peharz et~al.(2016)Peharz, Gens, Pernkopf, and Domingos]{peharz2016latent}
Robert Peharz, Robert Gens, Franz Pernkopf, and Pedro Domingos.
\newblock On the latent variable interpretation in sum-product networks.
\newblock \emph{IEEE transactions on pattern analysis and machine intelligence}, 39\penalty0 (10):\penalty0 2030--2044, 2016.

\bibitem[Peharz et~al.(2020)Peharz, Vergari, Stelzner, Molina, Shao, Trapp, Kersting, and Ghahramani]{peharz2020random}
Robert Peharz, Antonio Vergari, Karl Stelzner, Alejandro Molina, Xiaoting Shao, Martin Trapp, Kristian Kersting, and Zoubin Ghahramani.
\newblock Random sum-product networks: A simple and effective approach to probabilistic deep learning.
\newblock In \emph{Uncertainty in Artificial Intelligence}, pages 334--344. PMLR, 2020.

\bibitem[Pipatsrisawat and Darwiche(2008)]{pipatsrisawat2008new}
Knot Pipatsrisawat and Adnan Darwiche.
\newblock New compilation languages based on structured decomposability.
\newblock In \emph{AAAI}, volume~8, pages 517--522, 2008.

\bibitem[Ruan et~al.(2021)Ruan, Ullrich, Severo, Townsend, Khisti, Doucet, Makhzani, and Maddison]{ruan2021improving}
Yangjun Ruan, Karen Ullrich, Daniel~S Severo, James Townsend, Ashish Khisti, Arnaud Doucet, Alireza Makhzani, and Chris Maddison.
\newblock Improving lossless compression rates via monte carlo bits-back coding.
\newblock In \emph{International Conference on Machine Learning}, pages 9136--9147. PMLR, 2021.

\bibitem[S{\"a}rkk{\"a} and Svensson(2023)]{sarkka2023bayesian}
Simo S{\"a}rkk{\"a} and Lennart Svensson.
\newblock \emph{Bayesian filtering and smoothing}, volume~17.
\newblock Cambridge university press, 2023.

\bibitem[Shao et~al.(2020)Shao, Molina, Vergari, Stelzner, Peharz, Liebig, and Kersting]{shao2020conditional}
Xiaoting Shao, Alejandro Molina, Antonio Vergari, Karl Stelzner, Robert Peharz, Thomas Liebig, and Kristian Kersting.
\newblock Conditional sum-product networks: Imposing structure on deep probabilistic architectures.
\newblock In \emph{International Conference on Probabilistic Graphical Models}, pages 401--412. PMLR, 2020.

\bibitem[Shih et~al.(2021)Shih, Sadigh, and Ermon]{shih2021hyperspns}
Andy Shih, Dorsa Sadigh, and Stefano Ermon.
\newblock Hyperspns: Compact and expressive probabilistic circuits.
\newblock \emph{Advances in Neural Information Processing Systems}, 34, 2021.

\bibitem[Song et~al.(2020)Song, Sohl-Dickstein, Kingma, Kumar, Ermon, and Poole]{song2020score}
Yang Song, Jascha Sohl-Dickstein, Diederik~P Kingma, Abhishek Kumar, Stefano Ermon, and Ben Poole.
\newblock Score-based generative modeling through stochastic differential equations.
\newblock \emph{arXiv preprint arXiv:2011.13456}, 2020.

\bibitem[Tancik et~al.(2020)Tancik, Srinivasan, Mildenhall, Fridovich-Keil, Raghavan, Singhal, Ramamoorthi, Barron, and Ng]{tancik2020fourier}
Matthew Tancik, Pratul Srinivasan, Ben Mildenhall, Sara Fridovich-Keil, Nithin Raghavan, Utkarsh Singhal, Ravi Ramamoorthi, Jonathan Barron, and Ren Ng.
\newblock Fourier features let networks learn high frequency functions in low dimensional domains.
\newblock \emph{Advances in Neural Information Processing Systems}, 33:\penalty0 7537--7547, 2020.

\bibitem[Townsend et~al.(2019)Townsend, Bird, and Barber]{townsend2019practical}
James Townsend, Tom Bird, and David Barber.
\newblock Practical lossless compression with latent variables using bits back coding.
\newblock \emph{arXiv preprint arXiv:1901.04866}, 2019.

\bibitem[Tran et~al.(2019)Tran, Vafa, Agrawal, Dinh, and Poole]{tran2019discrete}
Dustin Tran, Keyon Vafa, Kumar Agrawal, Laurent Dinh, and Ben Poole.
\newblock Discrete flows: Invertible generative models of discrete data.
\newblock \emph{Advances in Neural Information Processing Systems}, 32, 2019.

\bibitem[Trapp et~al.(2019)Trapp, Peharz, Ge, Pernkopf, and Ghahramani]{trapp2019bayesian}
Martin Trapp, Robert Peharz, Hong Ge, Franz Pernkopf, and Zoubin Ghahramani.
\newblock Bayesian learning of sum-product networks.
\newblock \emph{arXiv preprint arXiv:1905.10884}, 2019.

\bibitem[Vergari et~al.(2019)Vergari, Di~Mauro, and Van~den Broeck]{vergari2019tractable}
Antonio Vergari, Nicola Di~Mauro, and Guy Van~den Broeck.
\newblock Tractable probabilistic models: {R}epresentations, algorithms, learning, and applications, 2019.
\newblock Tutorial at the 35th Conference on Uncertainty in Artificial Intelligence (UAI).

\bibitem[Vergari et~al.(2021)Vergari, Choi, Liu, Teso, and Van~den Broeck]{vergari2021compositional}
Antonio Vergari, YooJung Choi, Anji Liu, Stefano Teso, and Guy Van~den Broeck.
\newblock A compositional atlas of tractable circuit operations: From simple transformations to complex information-theoretic queries.
\newblock \emph{arXiv preprint arXiv:2102.06137}, 2021.

\bibitem[Xiao et~al.(2017)Xiao, Rasul, and Vollgraf]{xiao2017fashion}
Han Xiao, Kashif Rasul, and Roland Vollgraf.
\newblock Fashion-mnist: a novel image dataset for benchmarking machine learning algorithms.
\newblock \emph{arXiv preprint arXiv:1708.07747}, 2017.

\bibitem[Yang et~al.(2023)Yang, Gala, and Peharz]{yang2023bayesian}
Yang Yang, Gennaro Gala, and Robert Peharz.
\newblock Bayesian structure scores for probabilistic circuits.
\newblock In \emph{International Conference on Artificial Intelligence and Statistics}, pages 563--575. PMLR, 2023.

\bibitem[Yann et~al.(2010)Yann, Cortes, and CJ]{mnist}
LeCun Yann, Corinnam Cortes, and Burges CJ.
\newblock Mnist handwritten digit database, 2010.
\newblock URL \url{http://yann.lecun.com/exdb/mnist}.

\bibitem[Zhao et~al.(2016)Zhao, Poupart, and Gordon]{zhao2016unified}
Han Zhao, Pascal Poupart, and Geoff Gordon.
\newblock A unified approach for learning the parameters of sum-product networks.
\newblock \emph{arXiv preprint arXiv:1601.00318}, 2016.

\bibitem[Ziegler and Rush(2019)]{ziegler2019latent}
Zachary Ziegler and Alexander Rush.
\newblock Latent normalizing flows for discrete sequences.
\newblock In \emph{International Conference on Machine Learning}, pages 7673--7682. PMLR, 2019.

\end{thebibliography}


\newpage
\appendix
\onecolumn
\setcounter{section}{0}
\setcounter{table}{0}
\setcounter{figure}{0}
\setcounter{algorithm}{0}
\renewcommand{\thetable}{\Alph{section}.\arabic{table}}
\renewcommand{\thefigure}{\Alph{section}.\arabic{figure}}
\renewcommand{\thealgorithm}{\Alph{section}.\arabic{algorithm}}
\renewcommand{\thesection}{\Alph{section}}

\section{Algorithms}
\label{app:alg}

\cref{alg:bn2pic} shows how to convert an arbitrary Bayesian network with continuous LVs $\Z$
into a tree-shaped, smooth and structured-decomposable PIC via variable elimination \citep{darwiche2009modeling}.
\autoref{fig:lgm} shows the result of such algorithm when applied to a latent tree structure.

\begin{algorithm}[H]
\caption{BN2\PIC via latent variable elimination}
\label{alg:bn2pic}
\begin{flushleft}
\textbf{Input:} A Bayesian Network over observables $\rmX = \{X_i\}_{i=1}^D$ and continuous latents $\rmZ = \{Z_i\}_{i=1}^{D'}$ \\
\textbf{Output:} A tree-shaped, smooth and structured-decomposable \PIC
\end{flushleft}
\begin{algorithmic}[1]
\State $C \gets \textsc{SetOfUnits}\big( \inputunit \big(\big[p(X_i \cbar \pa(X_i))\big]\big)_{i=1}^D \big)$ \Comment{Initialise set of units with input units}
\vspace{0.5mm}
\For{$Z_i$ \textbf{in} $\textsc{ToList}(\rmZ)$} \Comment{Any order of $\rmZ$ is fine}
    \State $D \gets$ retrieve all units in $C$ whose output functions include $Z_i$ 
    \State $C \gets$ replace units $D$ with integral unit $\intu\big(\big[\!\bigotimes(D) \big] \big)$ having $p(Z_i \cbar \pa(Z_i))$ as attached density
\EndFor
\State $c \gets \bigotimes(C)$
\State $c \gets \textsc{OptionalPruning}(c)$
\State \Return $c$ \Comment{\PIC root unit}
\end{algorithmic}
\end{algorithm}

\cref{alg:quad} shows how to apply arbitrary quadrature rules (e.g.\ integrand-dependent) to approximate a \PIC into a \QPC whose size is exponential in the \PIC depth.
\autoref{fig:expqpc} shows the \QPC resulting from the application of \cref{alg:quad} on the \PIC in \autoref{fig:lgm}.
Note that our neural \PICs (cf.\ \cref{sec:learnpic}) are always materialised via Algorithm \ref{alg:pic2qpc}, therefore always delivering \QPCs whose sizes are quadratic in the number of integration points $N$.

\begin{algorithm}[!ht]
\caption{\textbf{\PIC $\rightarrow$ \QPC} via arbitrary nested univariate quadratures}
\label{alg:quad}
\begin{flushleft}
\textbf{Input:} A tree-shaped \PIC $\gC$ parameterised by $\{ p(X_i \cbar Z_i) \} \cup \{p(Z_i \cbar Z_{\pa(i)})\}$ with no sum units \\
\textbf{Output:} A \QPC representing an approximation of the input \PIC via nested univariate quadratures
\end{flushleft}
\begin{algorithmic}[1]
\vspace{1mm}
\State $c = $ \textsc{Quad}$(\textsc{GetRoot}(\gC), \textsc{null})$ \Comment{\PIC integral root unit has no conditioning RVs, thus $\tilde{z}=\textsc{null}$}
\State \Return $c$ \Comment{return \QPC sum root unit}
\vspace{1mm}
\Function{\textsc{Quad}}{$\intu$, $\tilde{z}$}
\State $\intz, \intw \gets \textsc{QuadPointsAndWeights}\big(p(Z_i \cbar Z_{\pa(i)} = \tilde{z})\big)$ \Comment{$(\intz, \intw)$ \emph{can} depend on the integrand} \label{ln:intchoice}
\State $R \gets \big[ \, \inputunit \, \big(p(X_i \cbar Z_i = \intz[n])\big) \big]_{n=1}^{\textsc{len}(\intz)}$
\If{$\intu$ has a product input unit}
    \State $C \gets \textsc{GetInputIntegralUnits}(\textsc{GetProdInputUnit}(\intu))$
    \State $R \gets \big[ \bigotimes \big( \, [R[n]] \cup [\textsc{Quad}(c, \intz[n])]_{c \in C} \big) \big]_{n=1}^{\textsc{len}(\intz)}$ \Comment{\emph{all} \textsc{Quad}$(\cdot)$ calls can run in parallel}
\EndIf
\State \Return $\bigoplus \big(R, \big[ \intw[n] \, p(Z_i = \intz[n] \cbar Z_{\pa(i)} = \tilde{z}) \big]_{n=1}^{\textsc{len}(\intz)} \big)$
\EndFunction
\end{algorithmic}
\end{algorithm}

\clearpage
\newpage
\section{Experiments}
\label{app:exp}

\paragraph{Training}

We train \PICs (resp.\ HCLTs) for a maximum of 30K training steps using a batch size of $B \in \{64, 128, 256\}$ and number of integration points (resp.\ categorical latent dimension) of $N \in \{16, 32, 64, 128\}$.
Adam learning rate and EM step size are both annealed from $10^{-2}$ to $10^{-4}$ using the \emph{cosine annealing with warm restarts} scheduler over a range of 500 training steps \citep{loshchilov2016sgdr}.
We use early stopping: Training stops when the average validation log-likelihood does not improve after 1250 training steps.
We train \PICs under the trapezoidal quadrature rule.
We use a single NVIDIA A100-SXM4-40GiB for all our experiments. \\ \\

\begin{figure}[H]
\centering
\begin{subfigure}{\textwidth}
  \centering
    \includegraphics[width=.75\textwidth]{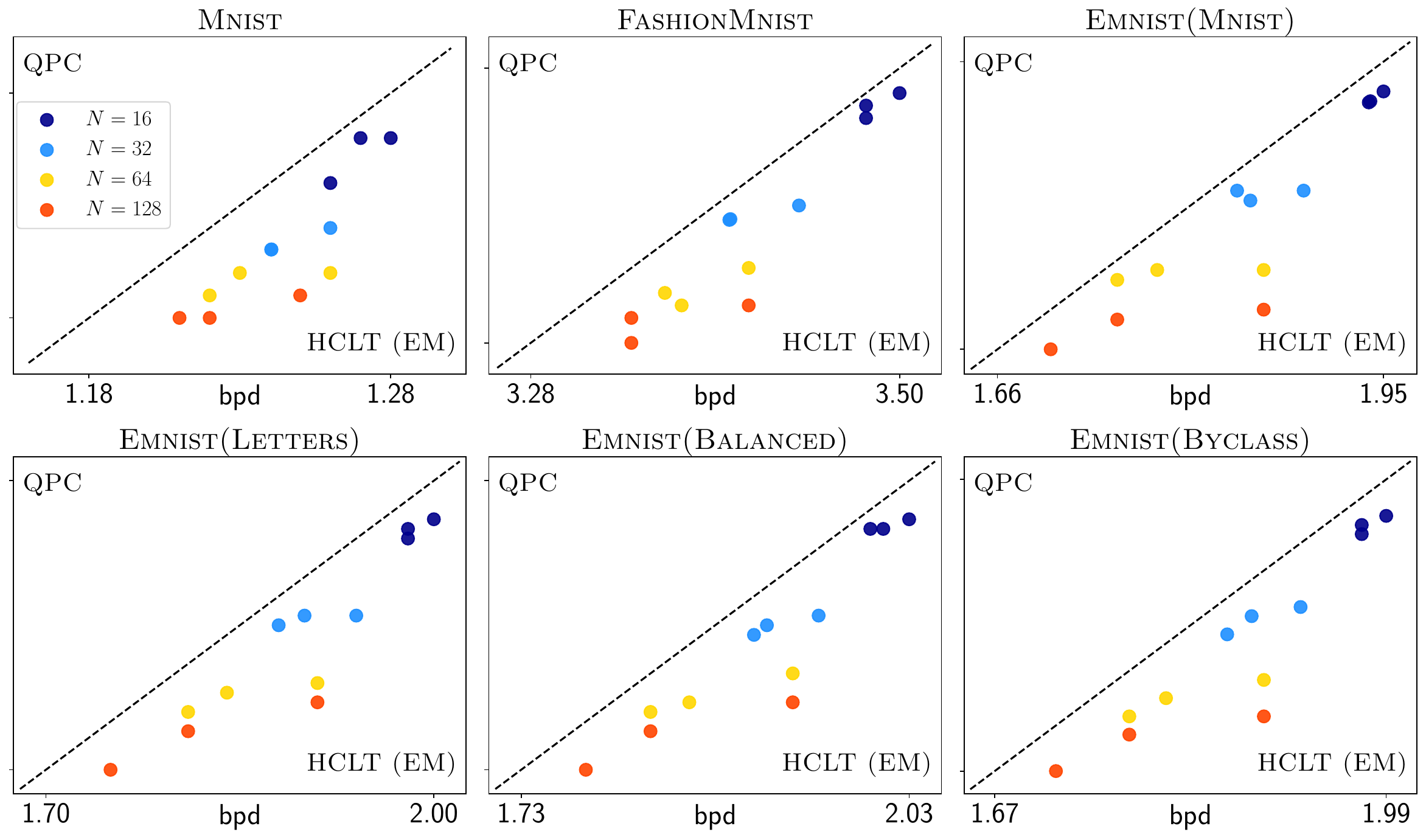}
  \caption{Categorical input units}
  \label{fig:sub1}
\end{subfigure}%
\vspace{\baselineskip}
\begin{subfigure}{\textwidth}
  \centering
    \includegraphics[width=.75\textwidth]{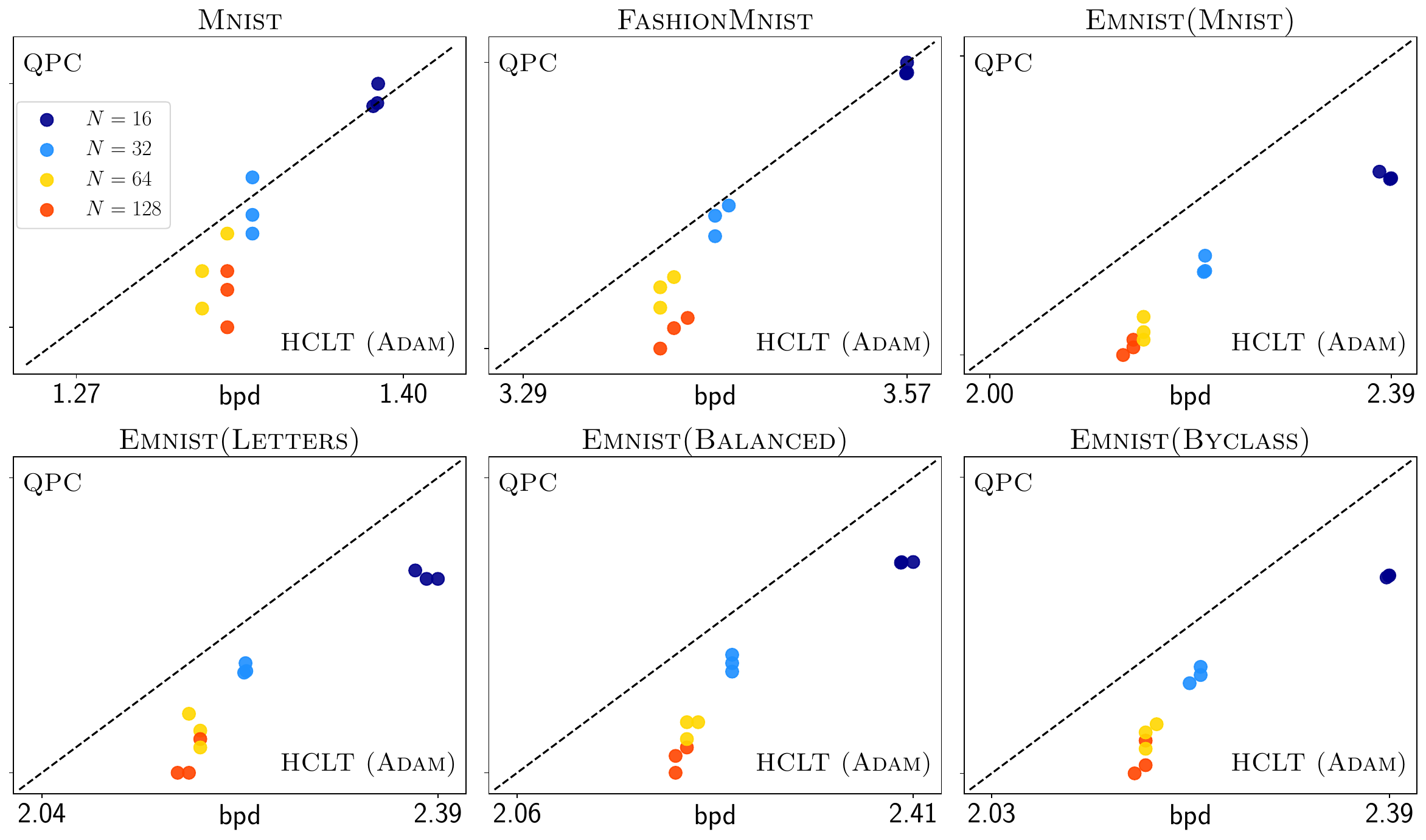}
  \caption{Binomial input units}
  \label{fig:sub2}
\end{subfigure}
\caption{
\textbf{\QPCs outperform HCLTS.}
Average test-set bits-per-dimension (bpd) of \QPCs (y-axis) and HCLTs (x-axis) paired by $N \in \{16, 32, 64, 128\}$ and batch size $B \in \{64, 128, 256\}$, for categorical (a) and binomial (b) input units.
In (a) HCLTs are trained via Expectation-Maximisation, whereas in (b) using Adam. 
\QPCs outperfom HCLTs in almost all hyperparameter configurations.
}
\label{fig:mnist_appendix}
\end{figure}

\subsection{Time \& Space}
\label{sec:timespace}

\begin{figure}[H]
  \centering
  \subfloat[Time (s)]{\includegraphics[width=0.45\textwidth]{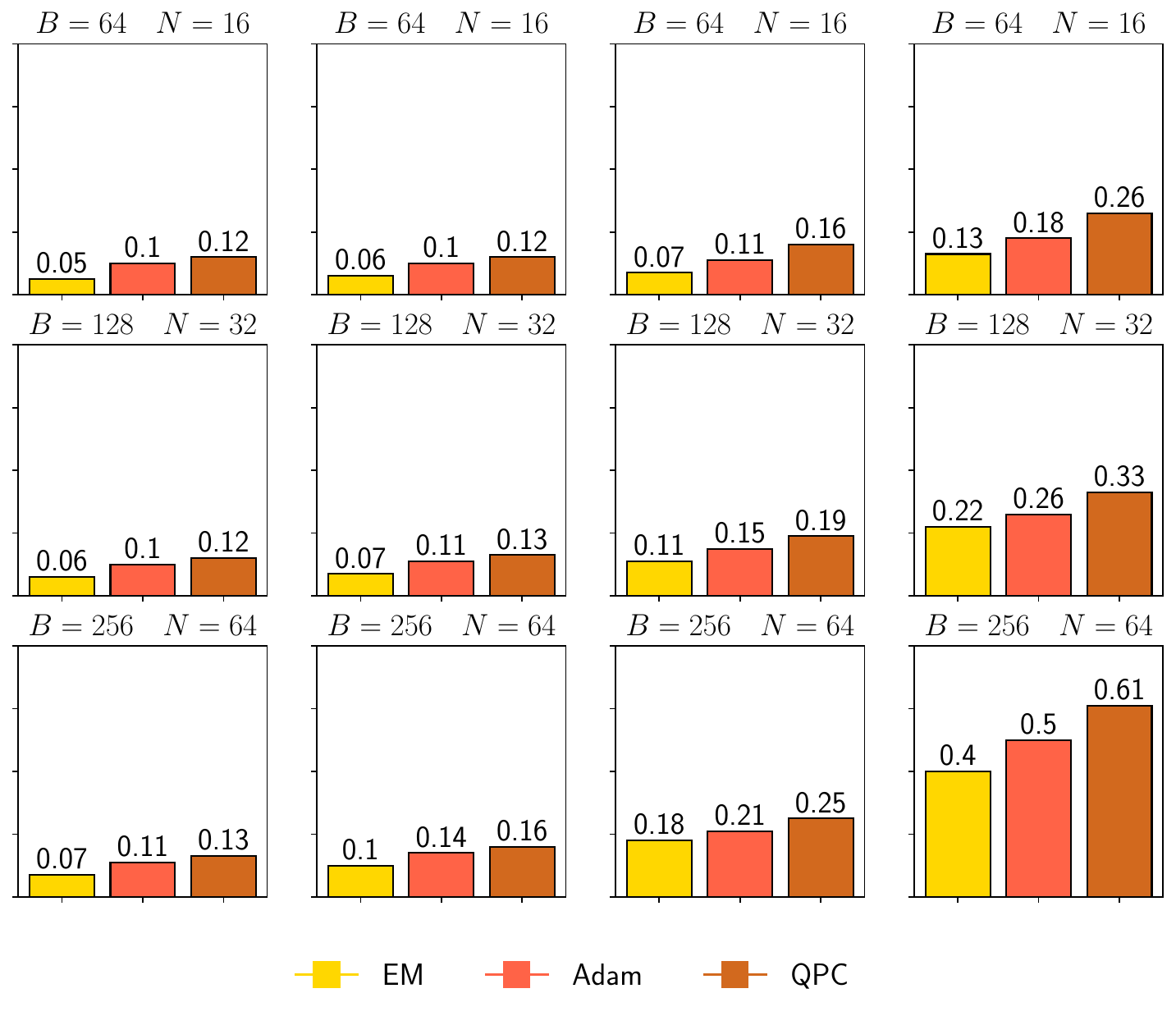}\label{fig:time}}
  \qquad
  \subfloat[GPU (GiB)]{\includegraphics[width=0.45\textwidth]{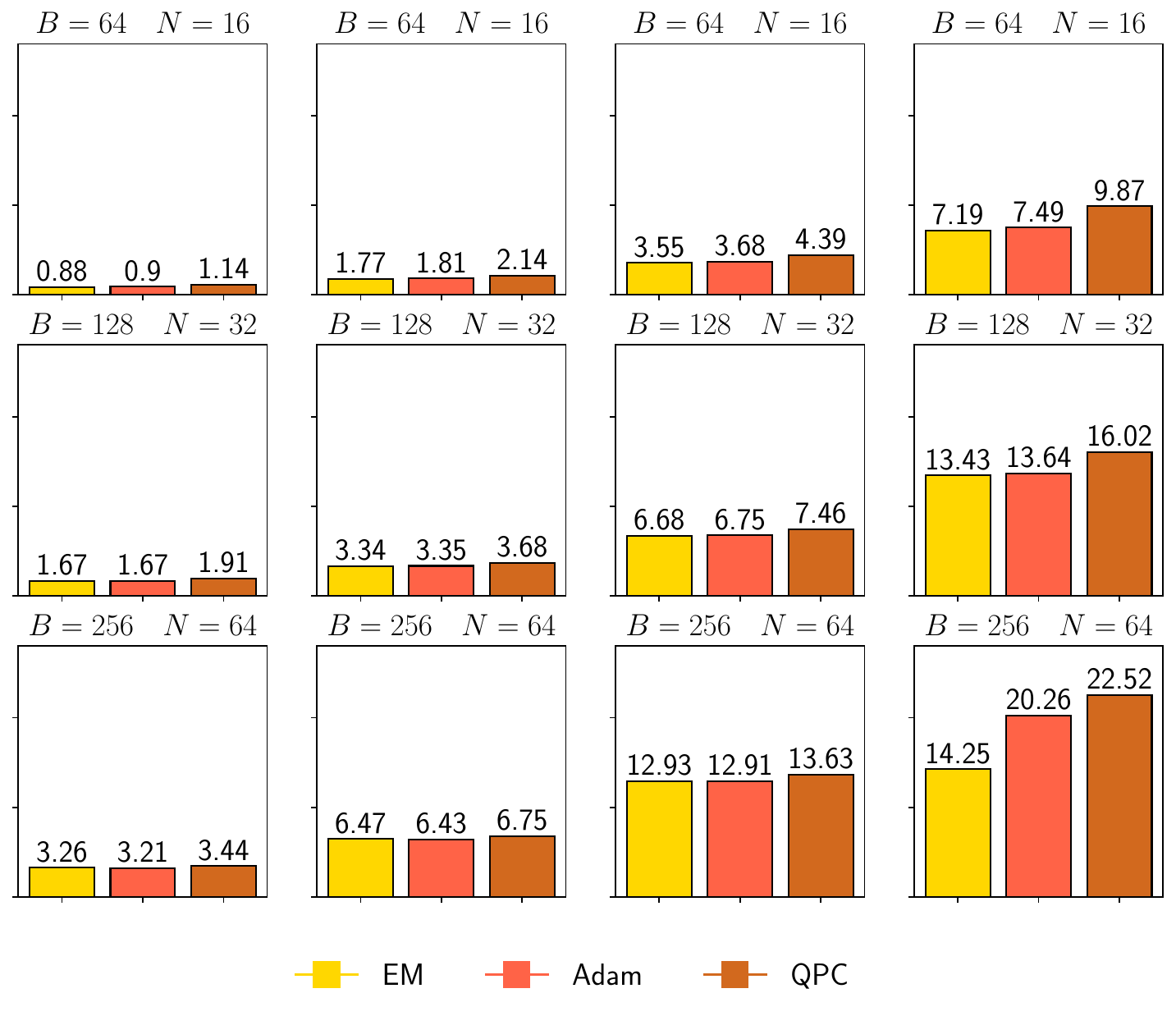}\label{fig:space}}
  \caption{
  \textbf{Training \PICs via numerical integration adds little time-space overhead.}
  We report time (a) and allocated GPU-memory peak (b) needed to process MNIST mini-batches of size $B \in \{64, 128, 256\}$ at different $N \in \{16, 32, 64, 128\}$ for HCLTs trained via EM (yellow) and Adam (orange), and \QPCs (brown).
  All PC models have categorical input units.
  }
  \label{fig:spacetime}
\end{figure}

In \autoref{fig:spacetime}, we report the time (in seconds) and space (in GiB) required to process MNIST mini-batches when \emph{training} under different learning schemes.
Although \PIC number of parameters can be significantly smaller than the number of materialised \QPC parameters, training \PICs via numerical integration (cf.\ \cref{sec:learnpic}) adds little time-space overhead due to the materialisation of \QPC parameters at each optimisation step.
Instead, at test-time, being \QPCs standard PCs, they require the same inference time and space of standard HCLTs, if materialised using Algorithm \ref{alg:pic2qpc}.

\subsection{Fourier Features vs No Fourier Features}
\label{sec:fourier_vs_no_fourier}

\begin{figure}[H]
\centering
\begin{subfigure}{0.4\textwidth}
  \centering
  \includegraphics[width=0.9\linewidth]{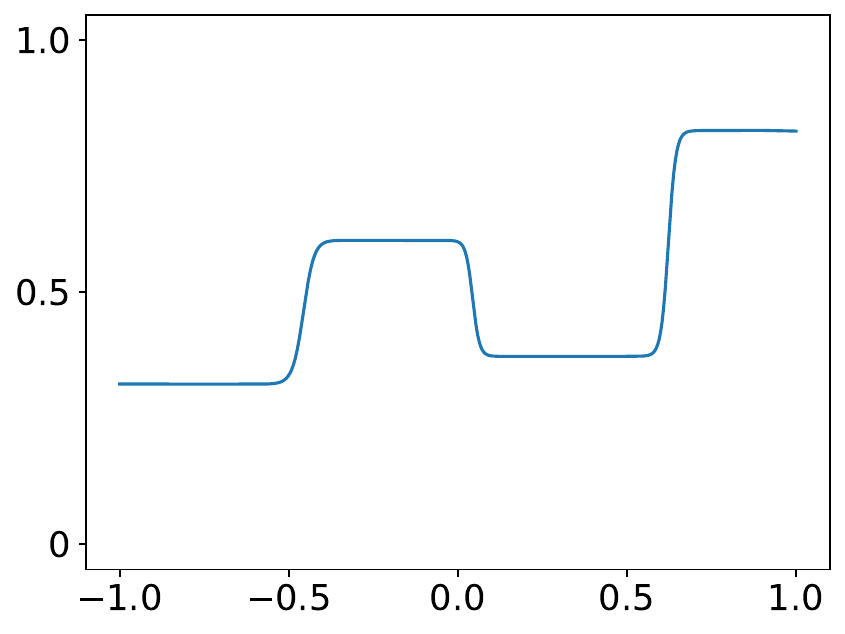}
  \caption{MLP}
  \label{fig:fourier}
\end{subfigure}
\begin{subfigure}{0.4\textwidth}
  \centering
  \includegraphics[width=0.9\linewidth]{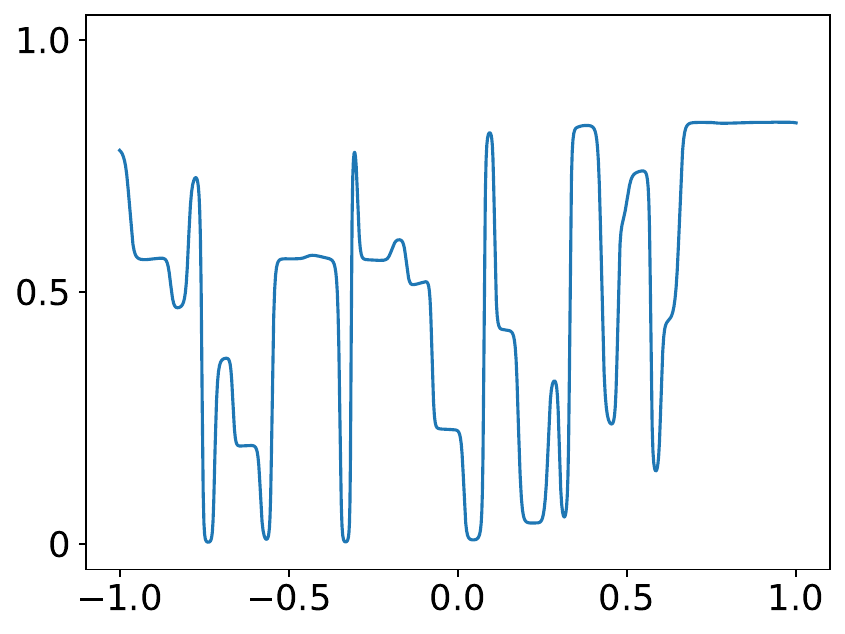}
  \caption{MLP with FFL}
  \label{fig:no_fourier}
\end{subfigure}
\caption{
\textbf{Fourier Features Layers allow modelling expressive distributions.}
\PIC root unit density learned using a standard MLP (a) vs.\ an MLP with FFL in (b).
}
\label{fig:fourier_vs_no_fourier}
\end{figure}

Fourier Feature Layers \citep{tancik2020fourier} are essential to get competitive performance with \PICs.
In fact, when using simple multi-layer perceptrons (MLPs) without FFLs, materialised \QPCs with categorical input units could not score better than 1.27 bpd on MNIST.
We argue that this is because FFLs allow modelling expressive distribution in contrast to possible oversmoothing behaviour of simple MLPs, as show in figure \autoref{fig:fourier_vs_no_fourier}.

\subsection{Sum region heatmaps}
\begin{figure}[H]
\centering
\begin{subfigure}{0.4\textwidth}
  \includegraphics[width=1.0\linewidth]{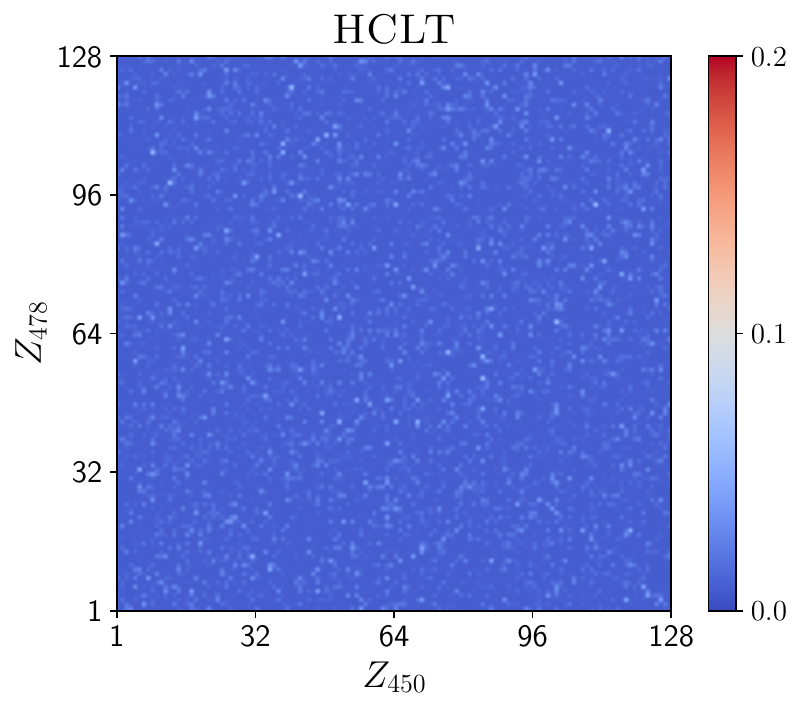}
  \caption{$\textsc{HCLT}$ sum region heatmap}
  \label{fig:hclt_heatmap}
\end{subfigure}\qquad
\begin{subfigure}{0.4\textwidth}
  \includegraphics[width=1.0\linewidth]{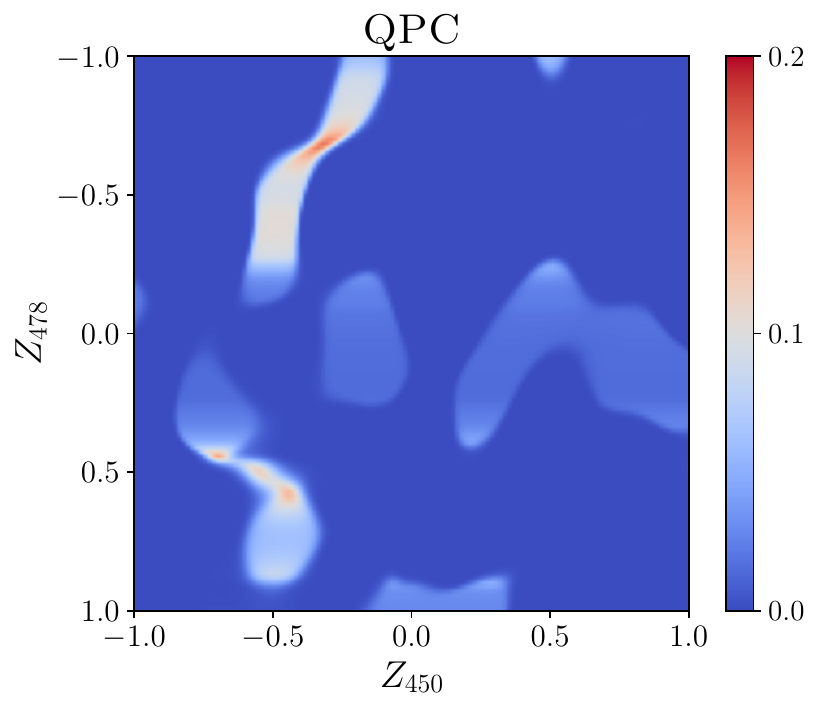}
  \caption{$\QPC$ sum region heatmap}
  \label{fig:qpc_heatmap}
\end{subfigure}
\caption{
\textbf{\QPC sum regions present structure, unlike PC ones.}
$Z_{450}$ sum region heatmap of a standard HCLT(a) vs.\ a \QPC (b) both with $N=128$.
}
\label{fig:heatmaps}
\end{figure}

In \autoref{fig:heatmaps}, we show the significant (but expected) difference between HCLT and \QPC sum regions.
We do this by visualising sum region parameters $\gS \in [0, 1]^{N \times N}$ as a heatmap with child and parent states reported on the x and y-axis respectively.
Sum regions of standard HCLTs look like the one in \autoref{fig:hclt_heatmap}, which resembles pure noise with no structure in it.
This reflects the fact that standard sum units represent discrete LVs of unordered categorical states.
Instead, \QPC sum regions show interesting smooth patterns, as the one showed in \autoref{fig:qpc_heatmap}.

\section{\PICs and \QPCs for Linear Gaussian Models}
\label{app:lgm}

\subsection{Tractable Gaussian \PICs}
\label{app:gaussianpic}

In this section, we will show the symbolic computation of a tractable PIC representing a linear Gaussian LTM.
Specifically, we will solve every integration problem by making use of the following result \citep{sarkka2023bayesian}: If $\rmZ \thicksim \gN(\boldsymbol{\mu}, \boldsymbol{\Sigma})$ and $\rmX \cbar \rmZ \thicksim \gN(\rmA\rmZ + \rvb, \boldsymbol{\Omega})$ then $\rmX \thicksim \gN(\rmA \boldsymbol{\mu} + \rvb, \rmA \boldsymbol{\Sigma} \rmA^{\top} + \boldsymbol{\Omega})$.

\begin{figure}[H]
\centering
\includegraphics[scale=0.5]{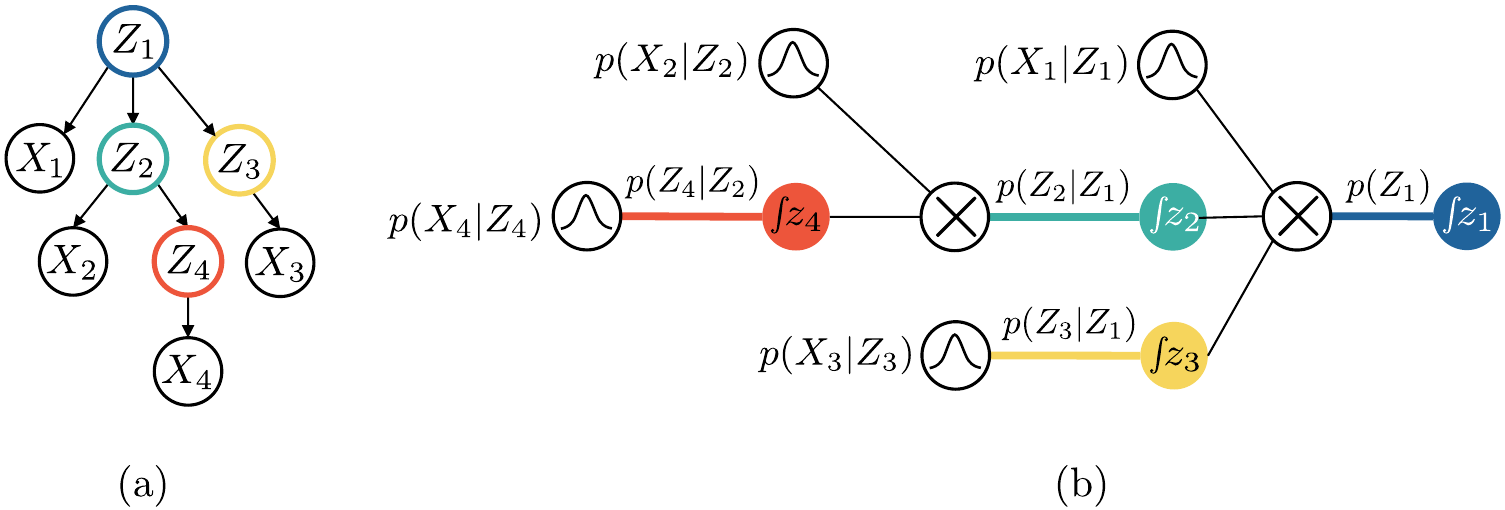}
\caption{An LTM (a) and its corresponding \PIC (b) compiled via Alg.\ \ref{alg:bn2pic} using $[Z_4, Z_2, Z_3, Z_1]$ as ordering.}
\label{fig:lgm}
\end{figure}

Suppose that the LTM structure in \autoref{fig:lgm}(a) is a linear-Gaussian LTM whose marginal over $\rmX$ is
\begin{equation*}
    p(\rmX) = \int \psi_1 (X_1, z_1) \int \psi_2 (X_2, z_2, z_1)  \int \psi_4 (X_4, z_4, z_2) dz_4 dz_2 \int \psi_3 (X_3, z_3, z_1) dz_3 dz_1,
\end{equation*}
where $\psi_i \equiv p(X_i \cbar Z_i) p(Z_i \cbar Z_{\pa(i)})$.
Since RVs are in linear relationships between them, we have that:
\begin{itemize}
    \item $Z_1 \sim \gN(\mu_1, \sigma_1^2)$
    \item $Z_i \sim \gN(a_i Z_{\pa(i)} + b_i, \sigma_i^2)$, where $i \neq 1$
    \item $X_i \sim \gN(c_i Z_i + d_i, \tau_i^2)$
\end{itemize}
where $a_i, b_i, c_i$ and $d_i$ belong to $\sR$.
Following the bottom-up computation of the \PIC in \autoref{fig:lgm}(b), we could evaluate \emph{in parallel} the integral unit $\intu[4]$ and $\intu[3]$ as:
\begin{align*}
\int \gN(z_4 \cbar a_4 z_2 + b_4, \sigma_4^2) \gN(X_4 \cbar c_4 z_4 + d_4, \tau_4^2) dz_4 = \gN(X_4 \cbar c_4(a_4 z_2 + b_4)+d_4, c_4^2\sigma_4^2 + \tau_4^2) \\
\int \gN(z_3 \cbar a_3 z_1 + b_3, \sigma_3^2) \gN(X_3 \cbar c_3 z_3 + d_3, \tau_3^2) dz_3 = \gN(X_3 \cbar c_3(a_3 z_1 + b_3)+d_3, c_3^2\sigma_3^2 + \tau_3^2)
\end{align*}
Setting $\tilde{c}_4 = c_4a_4$, $\tilde{d}_4 = c_4b_4 + d_4$, $\tilde{\tau}_4^2 = c_4^2\sigma_4^2 + \tau_4^2$, $\tilde{c}_3 = c_3a_3$, $\tilde{d}_3 = c_3b_3 + d_3$ and $\tilde{\tau}_3^2 = c_3^2\sigma_3^2 + \tau_3^2$, we rewrite the integrals above as
\begin{align*}
\int \gN(z_4 \cbar a_4 z_2 + b_4, \sigma_4^2) \gN(X_4 \cbar c_4 z_4 + d_4, \tau_4^2) dz_4 = \gN(X_4 \cbar \tilde{c_4}z_2 + \tilde{d_4}, \tilde{\tau}_4^2) \\
\int \gN(z_3 \cbar a_3 z_1 + b_3, \sigma_3^2) \gN(X_3 \cbar c_3 z_3 + d_3, \tau_3^2) dz_3  = \gN(X_3 \cbar \tilde{c_3}z_1 + \tilde{d_3}, \tilde{\tau}_3^2)
\end{align*}
Then, we evaluate the integral unit $\intu[2]$, using the result of unit $\intu[4]$ as
\begin{align*}
    \int \gN(z_2 \cbar a_2z_1 + b_2, \sigma_2^2) \gN(X_2 \cbar c_2z_2 + d_2, \tau_2^2) \gN(X_4 \cbar \tilde{c}_4 z_2 + \tilde{d}_4, \tilde{\tau}_4^2) dz_2 = \gN((X_2, X_4)^{\top} \cbar \boldsymbol{\mu}^{\prime}, \boldsymbol{\Sigma}^{\prime}).
\end{align*}
where
\begin{align*}
\boldsymbol{\mu}^{\prime} = 
\begin{pmatrix*}[l]
    c_2 (a_2 z_1 + b_2) + d_2 \\
    \tilde{c}_4 (a_2 z_1 + b_2) + \tilde{d}_4
\end{pmatrix*},
\quad
\boldsymbol{\Sigma}^{\prime} = 
\begin{pmatrix*}
    c_2^2 \sigma_2^2 + \tau_2^2, & \\
    c_2^2 \tilde{c}_4^2 \sigma_2^2, & \tilde{c}_4^2\sigma_2^2+\tilde{\tau}_4^2
\end{pmatrix*}.
\end{align*}
Finally, we evaluate $\intu[1]$---using the output functions from $\intu[2]$ and $\intu[3]$---as
\begin{align*}
    \int \gN(z_1 \cbar \mu_1, \sigma_1^2) \gN(X_1 \cbar c_1 z_1 + d_1, \tau_1^2) \gN((X_2, X_4)^{\top} \cbar \boldsymbol{\mu}^{\prime}, \boldsymbol{\Sigma}^{\prime}) \gN(X_3 \cbar \tilde{c}_3 z_1 + \tilde{d}_3, \tilde{\tau}_3^2) dz_1 = \gN(\rmX \cbar \boldsymbol{\mu}, \boldsymbol{\Sigma}),
\end{align*}
where
\begin{align*}
\boldsymbol{\mu} = 
\begin{pmatrix*}
    c_1\mu_1 + d_1 \\
    c_2(a_2\mu_1 + b_2) + d_2 \\
    \tilde{c_3}\mu_1 + \tilde{d_3} \\
    \tilde{c_4}(a_2\mu_1 + b_2) + \tilde{d_4}
\end{pmatrix*},
\end{align*}
\begin{align*}
\boldsymbol{\Sigma} = 
\begin{pmatrix*}
    c_1^2 \sigma_1^2 + \tau_1^2, &      \\
    a_1 a_2 c_2 \sigma_1^2, & a_2^2 c_2^2 \sigma_1^2 + c_2^2 \sigma_2^2+\tau_2^2     \\
    c_1 \tilde{c}_3 \sigma_1^2, &a_2 c_2 \tilde{c}_3 \sigma_1^2, &\tilde{c}_3^2\sigma_1^2 + \tilde{\tau}_3^2 \\
    a_2 c_1 \tilde{c}_4 \sigma_1^2, &a_2^2 c_2 \tilde{c}_4 \sigma_1^2 + c_2 \tilde{c}_4 \sigma_2^2, &a_2 \tilde{c}_3 \tilde{c}_4 \sigma_1^2, & a_2^2 \tilde{c}_4^2 \sigma_1^2 + \tilde{c}_4\sigma_2^2+\tilde{\tau}_4^2
\end{pmatrix*}.
\end{align*}

\subsection{Approximating linear-Gaussian models with \QPCs}
\label{app:sanitycheck}

In this section, we discuss details and insights behind our sanity check in \cref{sec:sanitycheck}.
Specifically, consider again the scenario in \cref{app:gaussianpic} and suppose we want to numerically approximate the \PIC in \autoref{fig:lgm}(b)---despite it being tractable---with a \QPC.

Different from our energy-based distributions (cf.\ \cref{sec:learnpic}), Gaussians have an infinite support, $\sR$.
Nonetheless, a Gaussian $\gN(\mu, \sigma^2)$ has a coverage of 99.7\% within 3 standard deviations, and therefore using $[\mu - 3\sigma, \, \mu+3\sigma]$ as integration domain could already be sufficient for a good numerical approximation, provided that the number of quadrature points $N$ is large enough.
However, the choice of such integration domain is integrand-dependent as the endpoints depend on the $\mu$ and $\sigma$ of the Gaussian we are integrating out.
As explained in \cref{sec:qpc} this is problematic.
Indeed, for every one of the $N$ integration points $\intz$ used to condition on a parent LV $Z_{\pa(i)}$, we instantiate $N$ distinct Gaussian distributions of the form $\gN(a_i \intz[n] + b_i, \sigma_i^2)$, with $n$ ranging from 1 to $N$.
Every sum unit in a \QPC is therefore associated with a specific Gaussian distribution, and thus requires specific integration domains.
As a consequence, the application of integrand-dependent quadrature rules deliver an exponentially big \QPC in the depth of the \PIC to approximate.

\begin{figure}[H]
\centering
\includegraphics[scale=0.46]{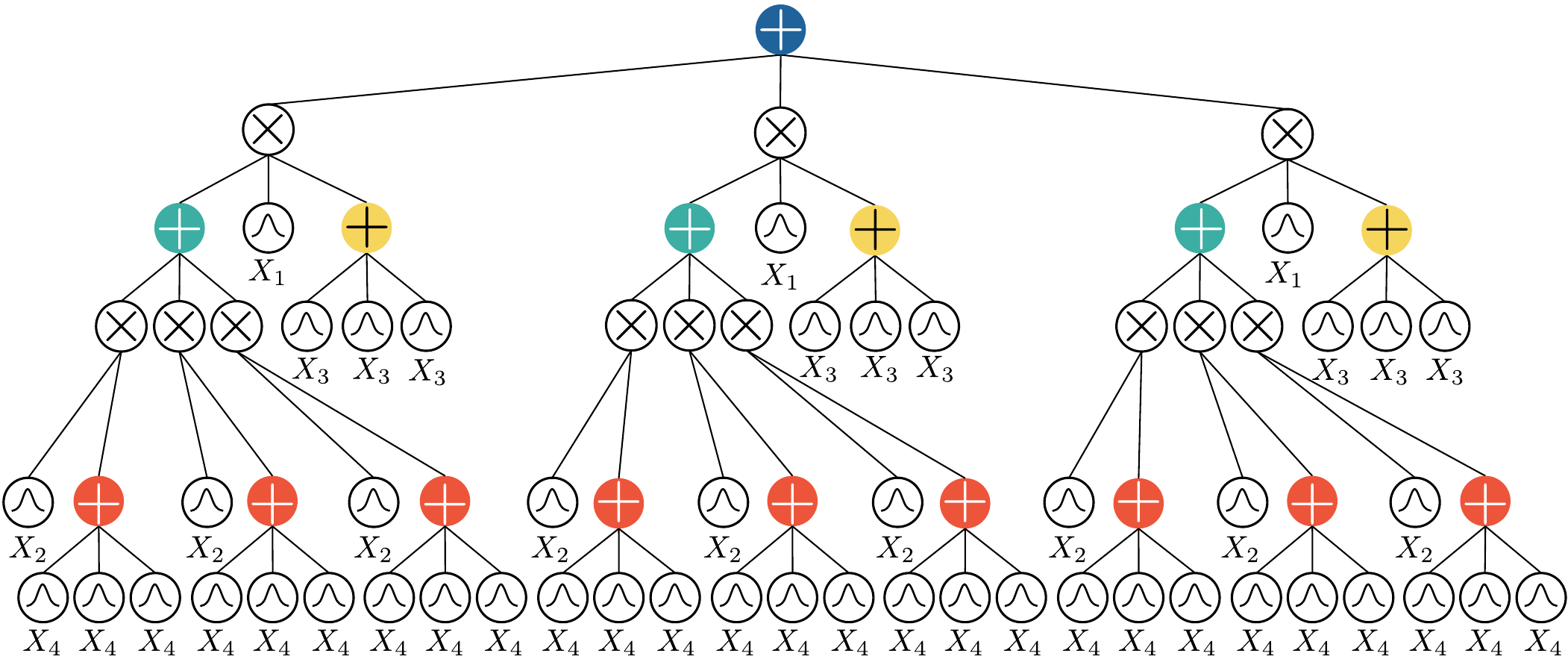}
\caption{
\textbf{Exponentially large \QPC via integrand-dependent quadrature rules.}
A \QPC materialised from the \PIC in \autoref{fig:lgm}(b)  via \cref{alg:quad} using integrand-dependent numerical quadrature rules.
The \QPC is a tree as unit re-usage is generally not possible when using integrand-dependent rules.
}
\label{fig:expqpc}
\end{figure}

As an example, applying integrand-dependent quadrature rules for approximating the \PIC in \autoref{fig:lgm}(b) could result in the \QPC shown in \autoref{fig:expqpc}, if using $N=3$ quadrature points for each integral approximation.
Specifically, such \QPC is obtained via \cref{alg:quad}, choosing the integration points in line \ref{ln:intchoice} within the interval $[\mu - 3\sigma, \mu+3\sigma]$ where $p(Z_i \cbar Z_{\pa(i)} = \tilde{z}) = \gN(\mu, \sigma^2)$.
The circuit has no unit re-usage as the integration points are chosen independently for each sum unit.
Note that, such exponential blow-up does \emph{not} arise neither because of the usage of Gaussian distributions nor for the infinite integration domain, but only for using integrand-dependent integration points.
As a consequence, we cannot represent such a nested quadrature approximation with a \QPC whose size is quadratic in the number of integration points $N$, as no re-usage of units is possible.

To approximate the \PIC in \autoref{fig:lgm}(b) using \cref{alg:pic2qpc} and therefore maintaining a quadratically large \QPC, we avoid integrand-dependent quadrature rules in favor of large integration domains, so as ensuring a good coverage for all the Gaussian sum units within a region.
Specifically, for each integral unit $\intu$ with density $p(Z_i | Z_{\pa(i)})$, we set the integration domain to $[\widecheck{\mu} - 3\sigma_i, \widehat{\mu} + 3\sigma_i]$, where $\widecheck{\mu} = \min_{n \in [N]} a_i \intz[n] + b_i$, $\widecheck{\mu} = \max_{n \in [N]} a_i \intz[n] + b_i$ and $\intz$ is the set of integration points used for $Z_{\pa(i)}$.
This is exactly what we do in \cref{sec:sanitycheck}.
This quadrature process proves effective, as shown in \autoref{fig:qpc_gauss}, but only when the coefficients of the linear transformations, $a_i$, fall within a relatively small range, e.g.\ $[-2, 2]$.
Instead, when arbitrary large coefficients are used, the integration bounds rapidly expand, rendering our static quadrature approach impractical.

\end{document}